\documentclass[10pt,journal,compsoc]{IEEEtran}

\usepackage{amsmath}
\usepackage{amssymb}
\usepackage{amsfonts}
\usepackage{amsthm}
\usepackage{graphicx}
\usepackage{mathtools}
\usepackage{balance}
\usepackage{bm}
\usepackage{colortbl}
\usepackage{lipsum}      
\usepackage{changepage}
\usepackage{caption}
\usepackage{color}
\usepackage{pgf,tikz}
\usepackage{balance}
\usepackage{mathtools}
\usepackage{bbm}
\usepackage{array}
\usepackage{relsize}
\usepackage{amsthm}
\usepackage{verbatim}
\usepackage{epstopdf}
\usepackage{array}
\usepackage{url}
\usepackage{stfloats}
\usepackage[hidelinks]{hyperref}
\usepackage{url}
\usepackage[linesnumbered,ruled]{algorithm2e}
\usepackage{algpseudocode}
\newtheorem{theorem}{Theorem}

\newtheorem{lemma}{Lemma}
\newtheorem{remark}{Remark}
\newtheorem{problem}{Problem}

\newtheorem{definition}{Definition}

\newcommand{\eqdef}{\mathrel{\mathop:}=}

\DeclareMathOperator*{\argmin}{arg\,min}
\usepackage{pdfpages} 

\ifCLASSOPTIONcompsoc
\usepackage[nocompress]{cite}
\else
\usepackage{cite}
\fi

\ifCLASSINFOpdf
\else

\fi

\begin{document}
	
	\title{Fundamental Limits of Deep Learning-Based Binary Classifiers Trained with Hinge Loss}

	\author{Tilahun~M.~Getu,~\IEEEmembership{Member,~IEEE,}
		Georges~Kaddoum,~\IEEEmembership{Senior Member,~IEEE,}
		and~Mehdi~Bennis,~\IEEEmembership{Fellow,~IEEE}
		\thanks{T. M. Getu is with the Electrical Engineering Department, \'Ecole de Technologie Sup\'erieure (ETS), Montr\'eal, QC H3C 1K3, Canada (e-mail: tilahun-melkamu.getu.1@ ens.etsmtl.ca).}

		\thanks{G. Kaddoum is with the Electrical Engineering Department, \'Ecole de Technologie Sup\'erieure (ETS), Montr\'eal, QC H3C 1K3, Canada (e-mail: georges.kaddoum@etsmtl.ca).}
		
		\thanks{M. Bennis is with the Centre for Wireless Communications, University of Oulu, 90570 Oulu, Finland (e-mail: mehdi.bennis@oulu.fi).}
		
	}

	
	\IEEEtitleabstractindextext{%
		\begin{abstract}
			Although deep learning (DL) has led to several breakthroughs in many disciplines, the fundamental understanding on why and how DL is empirically successful remains elusive. To attack this fundamental problem and unravel the mysteries behind DL's empirical successes, significant innovations toward a unified theory of DL have been made. Although these innovations encompass nearly fundamental advances in optimization, generalization, and approximation, no work has quantified the testing performance of a DL-based algorithm employed to solve a pattern classification problem. To overcome this fundamental challenge in part, this paper exposes the fundamental testing performance limits of DL-based binary classifiers trained with hinge loss. For binary classifiers that are based on deep rectified linear unit (ReLU) feedforward neural networks (FNNs) and deep FNNs with ReLU and Tanh activation, we derive their respective novel asymptotic testing performance limits, which are validated by extensive computer experiments.
		\end{abstract}
		
		\begin{IEEEkeywords}
			DL, the (unified) theory of DL, interpretable AI/ML, interpretable DL, fundamental limits.
		\end{IEEEkeywords}}

	\maketitle

	\IEEEdisplaynontitleabstractindextext

	%
	\IEEEpeerreviewmaketitle

\section{Introduction}
\label{sec: introduction}
\subsection{Related Works}
Being the most popular technology behind artificial intelligence (AI) / machine learning (ML), deep learning (DL) \cite{Lecun_DL_Nature_15} has led to numerous 21st-century breakthroughs in many disciplines. Among those disciplines, chemistry \cite{Protein_Struct_Pred'20,DNN_for_Drug_Discovery'15}; computer science \cite{mnih-DQN-2015,Mastering_Go_Nature'16,Tom_DL_Based_NLP'18}; electrical engineering \cite{Hinton_Speech_Recognition'12,CZPHH19,QFQH_18}; mathematics \cite{AI_Olympiad_2024,Han_Solving_High_Dim_PDEs'18}; medicine \cite{Skin_Cancer_Classification_Nat'17,DeFauw2018ClinicallyAD}; neuroscience \cite{Radhakrishnan_Overpara_FNN_Asso_Memory'20,Saxe_Math_Theory_Semantic_Development'19}; and physics \cite{Giuseppe_Quan_MB_ANNs,Quan_Entag_DL_Arch_19,DL_for_High_Energy_Physics_Baldi'14} have witnessed DL-based breakthroughs. Despite many breakthroughs, the fundamental understanding on \textit{why and how} DL is empirically successful in astronomically many AI/ML problems remains elusive. Nevertheless, it is widely believed that DL's empirical successes are attributable to its ability to learn salient hidden representations from the training data \cite{Bengio_Repre_Learning'13}, which goes through networked layers of interconnected \textit{closed-box processing} \cite{Castelvecchi_Black_Box_16} at the heart of DL's remarkable successes in solving both classification and regression AI/ML problems. The mysteries of this black box processing are also the mysteries behind DL's empirical achievements. To uncover these mysteries, the broader multidisciplinary AI/ML research communities have made significant theoretical advancements -- toward interpretable (or explainable) AI/ML -- that constitute \textit{the theory of DL} \cite{Poggio_Theo_Issues_Dnets_2020,Modern_DL_Math'21,Bartlett_DL_Stat_Viewpoint'21}, which is concerned with \textit{generalization, optimization, and approximation (representation) in DL} \cite{Poggio_Theo_Issues_Dnets_2020}. 
	
Inspired by the classical works \cite{Mhaskar_NN_Opt_Approx_96,Pinkus99approximationtheory} that have established that feedforward neural networks (FNNs) with one hidden layer are \textit{universal approximators}, approximation problems, in the theory of DL, deal with deriving the approximation error bounds of one or more function approximation(s) using deep networks. In this vein, a number of recent works have established that deep FNNs -- as opposed to shallow FNNs -- have the power of \textit{exponential expressivity} \cite{Rolnick_ICLR_18,Poole_Exponential_Expressivity'16,Lin_Deep_and_Cheap_Learning_2017,PMLR-v49-Eldan_16,Mhaskar_Deep_vs_Shallow_16,Poggio_When_and_How'17}. In light of exponential expressivity, the state-of-the-art literature includes many works on deep approximation such as the width efficiency of \textit{rectified linear unit} (ReLU) FNNs \cite{Lu_NIPS_Expressive_Power_of_NNs'17}; the quasi-equivalence of depth and width \cite{Fan_Dep_Wid_duality_20}; the universal approximation capability of slim (and sparse) networks \cite{Fan2018_Slim_and_sparse}; the approximation of univariate functions by deep ReLU FNNs \cite{Daubechies_NL_Approximation_and_DReLU_NNs'19}; approximation (and estimation) using high-dimensional DL networks \cite{Barron2018_DL_Approximation'18}; matrix-vector product approximation using deep ReLU FNNs \cite{Getu_error_bounds_nomal_21}; universal function approximation by deep ReLU FNNs \cite{Hanin_2019_Univ_Approx}; deep FNN approximation theory \cite{DL_Approx_Theory'21,Helmut_Optimal_Approximation'19}; and optimal approximation of piece-wise smooth functions using deep ReLU FNNs \cite{PP_optimal_approx_18}.
	
Apart from the aforementioned works on approximation in the context of the theory of DL, there also exist informative works (e.g., \cite{Geiger_DL_landscape_and_training'21,Sun_SPM_global_landscape'20}) on optimization in the context of the theory of DL. The authors of \cite{Baldassi_unrea_effectiveness_learning_NNs'16} highlight these works starting with useful developments for shallow FNNs and corroborate that there exist optimization landscape regions for discrete FNNs with only one or two layers that are both robust and accessible, and their existence is central to good training/testing performance. In addition, the authors of \cite{Baldassi_Shaping_WFM'20} derive various basic geometric and algorithmic features for nonconvex one- and two-layer FNNs that learn random patterns. These crucial advancements apply to shallow networks rather than the state-of-the-art practical deep networks that are often over-parameterized \cite{Sun_SPM_global_landscape'20}. Even though over-parameterized deep (and shallow) networks must be \textit{overfitted} per the classical bias–variance tradeoff \cite{Belkin_PNAS_reconciling_MML'19,NIPS2014_Comp_Efficiency}, over-fitting does not often happen with practical deep networks \cite{Sun_SPM_global_landscape'20}. This \textit{discrepancy}\footnote{Over-parameterized deep networks need to move only a tiny distance, and linearization can be a good approximation (also concerning the \textit{neural tangent kernel} (NTK) \cite{Jacot_NTK'18}). Nonetheless, because deep networks are not ultra-wide in practice, their parameters will move a large distance that is more than the linearization regimes permit \cite{Sun_SPM_global_landscape'20}. On the other hand, the authors of \cite{Chizat_lazy_training_2020} reveal an implicit bias phenomenon dubbed \textit{lazy training} that applies when a non-linear parametric model behaves like a linear one. This can occur implicitly under some choices of hyper-parameters governing normalization, initialization, and the number of iterations when the model is large \cite{Chizat_lazy_training_2020}.} remains as one of the fundamental issues in the theory of DL \cite{Sun_SPM_global_landscape'20}. Meanwhile, several theoretical works have been published on over-parameterized deep (and shallow) networks: Convergence theory for deep ReLU FNNs \cite{Allenzhu_convergence'18,Zou2018stochastic}; convergence theory for an over-parameterized ResNet \cite{Zhang2019_Convergence_Theory'19}; the optimization landscape of over-parameterized shallow FNNs \cite{Soltanolkotabi_TIT_19}; and how much over-parameterization is sufficient to learn deep ReLU FNNs \cite{Chen_overparameterization_2020}. Moreover, the recent theoretical works \cite{Allenzhu_Learing_and_Gen_2020,Cao_generalization_error_bounds_2019} affirm that over-parameterization improves generalization. In harmony with this theory, the authors of \cite{Zhang_Understanding'17} empirically corroborate that over-parameterized DL models can fit the training data and generalize well even when the labels are substituted by noise. Meanwhile, when it comes to generalization via implicit regularization and tractability via over-parametrization, the authors of \cite{Bartlett_DL_Stat_Viewpoint'21} provide a statistical viewpoint. 
	
As for generalization, state-of-the-art AI/ML research incorporates interesting works that exploit local Rademacher complexities, algorithmic stability, compression bounds, and algorithmic robustness to derive the generalization error bounds of deep (also shallow) networks \cite{Bartlett_DL_Stat_Viewpoint'21}. The authors of \cite{Belkin_PNAS_reconciling_MML'19}, on the other hand, expose the existence of \textit{double descent} that amalgamates the classical U-shaped test risk curve and the modern interpolating regime (with zero training error) for over-parameterized DL models. The authors of \cite{PNAS_citation_Adlam_Triple_Descent'20} took\linebreak inspiration from double descent and revealed \textit{triple descent}, which also comprises a superabundant parameterization regime corresponding to a deterministic NTK \cite{Jacot_NTK'18}. NTK is the limiting kernel of wide FNNs as their width approaches infinity \cite{Jacot_NTK'18}. When it comes to infinitely wide FNNs, crucial \textit{mean field limits} are derived for two-layer FNNs \cite{Mei_PNAS_NN_landscape_18} and multi-layer FNNs \cite{PNAS_citation_MF_Limit_Rrigorous_Framework'21}. Meanwhile, the fundamental phenomena of \textit{neural collapse} and \textit{minority collapse} are unveiled in \cite{Papyan_PNAS_prevalence_of_neural_collapse'20} and \cite{Fang_Minority_Collapse'21}, respectively. Neural collapse is a fundamentally pervasive inductive bias that occurs during the terminal phase of training beginning at the epoch in which the training error first vanishes while the training loss is driven to zero \cite{Papyan_PNAS_prevalence_of_neural_collapse'20}, whereas minority collapse is the fundamental limit on the minority classes of DL models that are trained for an adequately long time on class-imbalanced datasets \cite{Fang_Minority_Collapse'21}.
	
\subsection{Motivation and Context}
Foundational themes on the achievements of DL focus on the following questions: $i)$ How can training forward-propagating data through deep networks impact the optimization landscape to such an extent that viable solutions (such as \textit{good local minima}) are attained? $ii)$ If a viable solution is attained w.r.t. the training data, how can we generalize it well for unseen testing data? $iii)$ When it comes to learning using the \textit{back-propagation algorithm} \cite{Haykin_NNs_09} to solve both classification and regression AI/ML problems, how does a network trainer's choice of DL architecture, initializer, optimizer, normalization scheme, and regularization scheme interact with the training dataset(s) to influence the optimization landscape and the generalizability of a trained DL model? These questions must be carefully addressed so as to reveal the fundamental testing performance limits of a DL-based AI/ML algorithm. Therefore, quantifying the fundamental testing performance limits of a deep FNN-based AI/ML algorithm entails the quantification of the attainable optimization landscape during training and the subsequent generalization error bound.

There could be infinitely many local minima \cite{Understanding_ML_SSS'14} for an FNN architecture and a loss function, some of which can lead to efficient learning. However, we lack a fundamental insight on which minima the optimization algorithm selects due to a lack of: $i)$ understanding of training dynamics, and $ii)$ mathematical tools to describe the landscape of the loss function of large FNNs \cite{Weinan_E_Math_Under_NNB_ML'20}. Apart from these formidable challenges, the back-propagating gradients that are involved in error back-propagation are often random gradients with unknown distributions. These render the development of almost all \textit{high-dimensional statistics} \cite{wainwright_2019} nearly useless, especially when one is trying to determine the non-asymptotic testing performance limits of an FNN-based classification AI/ML algorithm.\footnote{For a regression AI/ML algorithm, the authors of \cite{Getu_Blind_DL_Estimator'22} are the first to provide asymptotic and non-asymptotic fundamental testing performance limits for a ReLU FNN-based regression AI/ML algorithm. The authors of \cite{Getu_TWC'24} derive the asymptotic performance limits on a DL-based text semantic communication system subjected to interference.} These fundamental testing performance limits may pave the way to the rigorous interpretation of trained deep networks solving classification AI/ML problems. To the best of our knowledge, nonetheless, no fundamental works that quantify the testing performance limits of DL-based classifiers -- trained with \textit{cross-entropy loss or hinge loss} -- have been managed yet. This is the motivation and context behind our work on the fundamental testing performance limits of FNN-based binary classifiers trained with hinge loss. 
	
\subsection{Contributions}
This paper investigates two kinds of DL-based binary classifiers: $i)$ The ones based on a deep ReLU FNN, and $ii)$ the ones based on a deep FNN with a ReLU activation function in all layers except the output layer equipped with a hyperbolic tangent activation function -- hereinafter referred to as a \textit{deep FNN with ReLU and Tanh activation}. We deliver the following major contributions:
	\begin{enumerate}
		\item We derive the fundamental asymptotic testing performance limits of binary classifiers based on a hinge-loss-trained deep ReLU FNN.
		
		\item We conduct extensive computer experiments whose results corroborate our derived asymptotic testing performance limits for deep ReLU FNN.
		
		\item We derive the asymptotic testing performance limits of binary classifiers based on a hinge-loss-trained deep FNN with ReLU and Tanh activation.
		
		\item We perform extensive computer experiments whose results validate our derived asymptotic testing performance limits for deep FNN with ReLU and Tanh activation.
	\end{enumerate}      
	
The remainder of this paper is organized as follows. Section \ref{sec: Prelude_and_System_Setup} outlines the prelude and system setup. Section \ref{sec: testing_performance_deep_FNN-Based_classifiers}\linebreak presents the problem formulations. Section \ref{sec: asymptotic_performance_limits_FNN_Based_binary_classifiers} documents the derived asymptotic testing performance limits. Section \ref{sec: results}\linebreak reports on extensive computer experiments. Last, Section \ref{sec: Conc_remarks_and_outlook} underscores our concluding remarks and research outlook. 
	
\textit{Notation}: Scalars, vectors, and matrices are represented by italic, bold lowercase, and bold uppercase letters, respectively. Uppercase italic letters denote random variables. Uppercase calligraphic letters denote events, hypotheses, and sets. $\mathbb{N}$, $\mathbb{N}_0$, $(\mathbb{R}_{+})\mathbb{R}$, $\mathbb{R}\backslash \{0\}$, $\mathbb{R}^{n}$, and $(\mathbb{R}_+^{m\times n})\mathbb{R}^{m\times n}$ denote the sets of natural numbers, natural numbers including zero, (positive)real numbers, real numbers excluding zero, $n$-dimensional real vectors, and $m\times n$ (positive)real matrices, respectively. $\sim$, $\eqdef$, $\equiv$, $\bm{0}$, $\bm{I}_n$, $\exp(\cdot)$, and $\mathbb{P}(\cdot)$ stand for distributed as, equal by definition, equivalent to, a zero vector/matrix whose dimension will be clear in context, an $n\times n$ identity matrix, exponential function, and probability, respectively. $\textnormal{diag}(\cdot)$, $\textnormal{sgn}(\cdot)$, $\textnormal{max}(\cdot, \ldots, \cdot)$, $\textnormal{min}(\cdot, \ldots, \cdot)$, $\mathcal{N}(\mu, \sigma^2)$, $(\cdot)^\top$, $\mathbb{I}\{\cdot\}$, and $|\cdot|$ stand for a diagonal matrix, the sign function, maximum of, minimum of, Gaussian distribution with mean $\mu$ and variance $\sigma^2$, transpose, a (component-wise) indicator function that returns 1 if the argument is true and 0 otherwise, and cardinality, respectively. 
	
	$\mathcal{Q}(\cdot)$ denotes the $\mathcal{Q}$-function defined for $x\in\mathbb{R}$ as $\mathcal{Q}(x) \eqdef \frac{1}{\sqrt{2\pi}}\int_{x}^{\infty} \exp(-t^2/2)dt$. $\argmin$, $\cap$, $\cup$, $\|\cdot \|$, $\delta(\cdot)$, and $\ln(\cdot)$ stand for the arguments of the minima, the intersection of two sets/events, the union of two sets/events, a vector norm, the Dirac delta function, and the natural logarithm, respectively. $\mathbb{N}_{\geq k}\eqdef \{k, k+1, \ldots \}$ denotes the sets of natural numbers greater than or equal to $k$ with $k\in \mathbb{N}_0$. For $n\in\mathbb{N}_{\geq 1}$, $[n]\eqdef\{1, \ldots, n\}$. For $\bm{A}\in\mathbb{R}^{m\times n}$ and $\bm{a}\in\mathbb{R}^{n}$, $(\bm{A})_{i,j}$ and $(\bm{a})_i$ denote the $(i,j)$-th element of $\bm{A}$ and the $i$-th entry of $\bm{a}$, respectively. Per the MATLAB\textsuperscript{\textregistered} syntax, $\bm{A}(i,:)$ denotes the $i$-th row of $\bm{A}\in\mathbb{R}^{m\times n}$. For $\bm{A}\in\mathbb{R}^{m\times n}$ and $\bm{a}\in\mathbb{R}^{n}$, $\|\bm{A}\|_{\ell_0}\eqdef |{(i, j)}: (\bm{A})_{i,j} \neq 0 |$, $\|\bm{a}\|_\infty\eqdef \max_{i=1, \ldots, n} |(\bm{a})_i|$, and $\|\bm{A}\|_\infty\eqdef \max_{i, j} |(\bm{A})_{i,j}|$. For $n\in\mathbb{N}_{\geq 2}$, the horizontal concatenation of $n$ conformable vectors and matrices is written as $[\bm{a}_1 \hspace{1mm} \bm{a}_2 \ldots \bm{a}_n]$ and $[\bm{A}_1 \hspace{1mm} \bm{A}_2 \ldots \bm{A}_n ]$, respectively.

\section{Prelude and System Setup}
\label{sec: Prelude_and_System_Setup}   
\subsection{Prelude}
\label{subsec: Prelude}    
\begin{definition}[{\textbf{Definition of FNNs \cite[Definition 2.2]{Ingo_Error_Bounds'19}}}]
	\label{def: NN_eq_definition}
	Let $K, N_0, N_1, \ldots, N_K\in\mathbb{N}$ and $K\geq 2$. A FNN $\bm{\Phi}$ is a sequence of matrix-vector tuples defined as
	\begin{equation}
		\label{NN_eq_def}
		\bm{\Phi}\eqdef\big[ [\bm{W}_1, \bm{b}_1], \bm{W}_2, \bm{b}_2],, \ldots, [\bm{W}_K, \bm{b}_K] \big], 
	\end{equation}
	where $\bm{W}_k\in\mathbb{R}^{N_k\times N_{k-1}}$ and $\bm{b}_k\in\mathbb{R}^{N_k}$ are the weight matrix of the neuronal connections between the $k$-th and $(k-1)$-th layers and the $k$-th layer's biases given $N_k$ is the number of neurons in the $k$-th layer (the input layer is considered the $0$-th layer), respectively. For the $K$-layer FNN expressed by (\ref{NN_eq_def}), $\mathcal{N}(\bm{\Phi})\eqdef\sum_{k=0}^K N_k$, $\mathcal{M}(\bm{\Phi})\eqdef\sum_{k=1}^{K} (\|\bm{W}_k\|_{\ell_0}+\|\bm{b}_k\|_{\ell_0})$, $\mathcal{W}(\bm{\Phi})\eqdef \max_{k=0, \ldots, K} N_k$, and $\mathcal{B}(\bm{\Phi})\eqdef \max_{k\in [K] } \big\{ \|\bm{W}_k\|_{\infty}, \|\bm{b}_k\|_{\infty}\big\}$ define the total number of neurons, the network connectivity, the maximum width, and the maximum absolute value of the weights, respectively. For an input $\bm{x}\in\mathbb{R}^{N_0}$ and an element-wise activation function $\rho(\cdot)$, an FNN map $\bm{\Phi}: \mathbb{R}^{N_0}\to\mathbb{R}^{N_K}$ is defined as $\bm{\Phi}(\bm{x})=\bm{x}_K$, where $\bm{x}_K$ is obtained recursively as   
	\begin{subequations}
		\begin{align}
			\label{NN_recur_1}
			\bm{x}_0&\eqdef \bm{x}; \hspace{2mm} \bm{x}_k\eqdef\rho (\bm{W}_k\bm{x}_{k-1}+\bm{b}_k), \hspace{2mm} \forall k\in [K-1]    \\
			\label{NN_recur_2}
			\bm{x}_K&\eqdef \bm{W}_K\bm{x}_{K-1}+\bm{b}_K,	 
		\end{align}
	\end{subequations}     
	where $\rho(\bm{y})=[\rho(y_1), \ldots, \rho(y_m)]^\top$ for $\bm{y}=[y_1, \ldots, y_m]^\top\in\mathbb{R}^{m}$. For $\rho(y_i)\eqdef\textnormal{max}(y_i, 0)$, the FNN in (\ref{NN_eq_def}) is known as a ReLU FNN. This type of FNN is often termed a deep ReLU neural network when $K$ is very large. When $N_k$, $k\in[K]$, is very large (often a polynomial order of the input number of neurons and the data points \cite{Allenzhu_convergence'18,Zou2018stochastic}), the FNN is usually termed an over-parameterized ReLU neural network if $K$ is small and an over-parameterized deep ReLU neural network if $K$ is very large.     
\end{definition}
\begin{definition}[{\textbf{FNNs with dual activation functions \cite{Haykin_NNs_09}}}]
	\label{ReLU_and_linear_FNNs_defn}
	Let $t\in\mathbb{N}_0$, $n\in\mathbb{N}$, and $K\in\mathbb{N}_{\geq 2}$. Let $N_1, N_k, N_{k-1}\in\mathbb{N}_{\geq 2}$ for $k\in[K]$. Assume an FNN training over $\mathcal{D}\eqdef \big\{ (x_n, y_n) \big\}_{n=1}^{N}$ given $x_n$ and $y_n\in\{-1, 1\}$ are the $n$-th input generated per (\ref{x_i__binary_class_problem}) and the $n$-th label, respectively. For a $K$-layer FNN with dual activation functions -- denoted by $\varphi(\cdot)$ and $\phi(\cdot)$ -- and $N_k$ neurons in its $k$-th layer, the network prediction during the $t$-th iteration is obtained as\footnote{As the biases can be subsumed by a linear model expressed by a non-zero weight matrix and an input equal to 1, they are discarded \cite{Haykin_NNs_09}.}    
	\begin{subequations}
		\begin{align}
			\label{y_n_1}
			\bm{y}_{1,n}^{(t)}&= \varphi(\bm{w}_1^{(t)}x_n)   \\
			\label{y_n_k}
			\bm{y}_{k,n}^{(t)}&= \varphi(\bm{W}_k^{(t)}\bm{y}_{k-1,n}^{(t)}), \hspace{2mm} \forall k\in\{2, \ldots, K-1\}  \\
			\label{y_n_K}
			y_{K,n}^{(t)}&= \phi(\bm{w}_K^{(t)}\bm{y}_{K-1,n}^{(t)}), 
		\end{align}
	\end{subequations}
	where $\varphi(\bm{x})=[\varphi(x_1), \ldots, \varphi(x_m)]^\top\in\mathbb{R}^m$ for $\bm{x}=[x_1, \ldots, x_m]^\top\in\mathbb{R}^m$; $\bm{w}_1^{(t)}\in\mathbb{R}^{N_1}$,\linebreak $\bm{W}_k^{(t)}\in\mathbb{R}^{N_k\times N_{k-1}}$, and $\big(\bm{w}_K^{(t)}\big)^\top\in\mathbb{R}^{N_{K-1}}$; $\bm{y}_{1,n}^{(t)}\in\mathbb{R}^{N_1}$, $\bm{y}_{k,n}^{(t)}\in\mathbb{R}^{N_k}$, and $y_{K,n}^{(t)}\in\mathbb{R}$ are the first, $k$-th, and $K$-th layers' outputs, respectively, at the $t$-th stochastic gradient descent (\texttt{SGD}) iteration\footnote{The $t$-th iteration accounts for the forward propagation of all the inputs of a mini-batch randomly chosen prior to forward-propagation.} in response to the $n$-th training input drawn from $\mathcal{D}$; and $\phi(\bm{y})=[\varphi(y_1), \ldots, \varphi(y_n)]^\top\in\mathbb{R}^n$ for $\bm{y}=[y_1, \ldots, y_n]^\top\in\mathbb{R}^n$.  
\end{definition}

\begin{remark}
For a ReLU FNN, $\varphi(x)=\rho(x)\eqdef \textnormal{max}(0,x)$ and $\phi(x)=x$ for $x\in\mathbb{R}$. For an FNN with ReLU and Tanh activation, $\varphi(x)=\rho(x)$ and $\phi(x)=\tanh(x)$ for $x\in\mathbb{R}$.	  
\end{remark}
\noindent We now proceed with a generic DL-based binary classification problem to investigate the fundamental limits of DL-based classification algorithms as applied to wireless communications and networking, signal processing, neuroscience, and quantum communications and networking.

\subsection{System Setup}
\label{subsec: System_Setup}
Consider a generic DL-based binary classification problem on hypotheses $\mathcal{H}_{-1}$ and $\mathcal{H}_1$ regarding the input datum $x_n$ defined through binary hypothesis testing as     
\begin{equation}
	\label{x_i__binary_class_problem}
	x_n=\begin{cases}
		f(\xi_n)+z_n &: \mathcal{H}_{-1} \\
		f(\beta_n)+z_n & : \mathcal{H}_1,  
	\end{cases}
\end{equation}
where $n\in\mathbb{N}$ and $x_n\in\mathbb{R}$; $f$ is an unknown generic function whose noisy outputs give rise to $x_n$; $\xi_n, \beta_n\in\mathbb{R}$ for $\xi_n \neq \beta_n$ are unknown data-generating input variables that are independent of $z_n\in\mathbb{R}$; and $z_n\sim\mathcal{N}(0, \sigma^2)$ is additive white Gaussian noise (AWGN) with zero mean and variance. To solve the binary pattern classification problem in (\ref{x_i__binary_class_problem}), we consider a deep FNN-based binary classifier trained using a training set $\mathcal{D}\eqdef \big\{ (x_n, y_n) \big\}_{n=1}^N$ -- given $x_n$ is the $n$-th input generated via (\ref{x_i__binary_class_problem}) and $y_n\in\{-1, 1\}$ is the $n$-th label -- and hinge loss per the back-propagation algorithm.\footnote{The derivation of the back-propagation algorithm (using square loss) is often documented using scalar calculus for an FNN equipped with the same activation function throughout all its layers \cite[Ch. 4]{Haykin_NNs_09}. For completeness and clarity, we employ \textit{matrix calculus} to derive the update rules for back-propagation w.r.t. hinge loss and \texttt{SGD} for FNNs with dual activation functions, as detailed in \cite[Appendix A, p. 45-52]{arXiv_Getu_Fundamental_Limits'23}.}

Let $(\bm{w}_K^{(t)})^\top\in\mathbb{R}^{2H}$, $\bm{W}_{K-k}^{(t)}\in\mathbb{R}^{2H\times 2H}$ $\forall k\in[K-2]$,\linebreak and $\bm{w}_1^{(t)}\in\mathbb{R}^{2H}$ be the weight matrices of the $K$-th layer, the $k$-th layer, and the first layer of a \textit{biasless} deep ReLU FNN under training during the $t$-th\linebreak iteration, respectively. During this iteration, $\bm{\Phi}^{(t)}\eqdef\big[ [\bm{w}_1^{(t)}, \bm{0}], [\bm{W}_2^{(t)}, \bm{0}],  \ldots,[\bm{W}_{K-1}^{(t)}, \bm{0}], [\bm{w}_K^{(t)}, \bm{0}] \big]$ defines a deep ReLU FNN under training and whose output is given by
\begin{equation}
	\label{y_n_K_simplified}
	\bm{\Phi}^{(t)}(x_n) \eqdef \bm{w}_K^{(t)} \bm{y}_{K-1,n}^{(t)} \in\mathbb{R}, 
\end{equation}
where, for $\bm{\Sigma}_{1,n}^{(t)}=\textnormal{diag}\big(\mathbb{I}\{\bm{w}_1^{(t)}x_n\geq 0\}\big)\in\{0,1\}^{2H\times 2H}$ and $\bm{\Sigma}_{k,n}^{(t)}=\textnormal{diag}\big(\mathbb{I}\big\{\bm{W}_k^{(t)}\big(\prod_{l=1}^{k-2}  \bm{\Sigma}_{k-l,n}^{(t)}\bm{W}_{k-l}^{(t)}\big)\bm{\Sigma}_{1,n}^{(t)}\bm{w}_1^{(t)}x_n\geq 0\big\}\big) \in\{0,1\}^{2H\times 2H}$ \cite{Zou2018stochastic},
\begin{equation}
	\label{y_n_K-1_simplified}
	\bm{y}_{K-1,n}^{(t)}=\Big(\prod_{k=1}^{K-2} \bm{\Sigma}_{K-k,n}^{(t)} \bm{W}_{K-k}^{(t)} \Big) \bm{\Sigma}_{1,n}^{(t)} \bm{w}_1^{(t)}x_n \in\mathbb{R}_{+}^{2H}.
\end{equation}
Note that $\bm{y}_{K-1,n}^{(t)}$ is the output of the $(K-1)$-th deep ReLU FNN layer during the $t$-th iteration. 

The output of a deep FNN  with ReLU and Tanh activation -- denoted by $\check{\bm{\Phi}}^{(t)}$ -- is equated as
\begin{equation}
	\label{y_n_K_simplified_with_tanh}
	\check{\bm{\Phi}}^{(t)}(x_n) \eqdef \tanh(\check{\bm{w}}_K^{(t)} \check{\bm{y}}_{K-1,n}^{(t)}) \in\mathbb{R}, 
\end{equation} 
where $\check{\bm{\Phi}}^{(t)}\eqdef\big[ [\check{\bm{w}}_1^{(t)}, \bm{0}], [\check{\bm{W}}_2^{(t)}, \bm{0}],  \ldots,[\check{\bm{W}}_{K-1}^{(t)}, \bm{0}], [\check{\bm{w}}_K^{(t)}, \bm{0}] \big]$ given $(\check{\bm{w}}_K^{(t)})^\top\in\mathbb{R}^{2H}$, $\check{\bm{W}}_{K-k}^{(t)}\in\mathbb{R}^{2H\times 2H}$ for all $k\in[K-2]$, and $\check{\bm{w}}_1^{(t)}\in\mathbb{R}^{2H}$, and $\check{\bm{y}}_{K-1,n}^{(t)}$ is the output of the $(K-1)$-th deep FNN layer during the $t$-th iteration, which is given by
\begin{equation}
	\label{y_n_K-1_simplified_with_tanh}
	\check{\bm{y}}_{K-1,n}^{(t)}=\Big(\prod_{k=1}^{K-2} \check{\bm{\Sigma}}_{K-k,n}^{(t)} \check{\bm{W}}_{K-k}^{(t)} \Big) \check{\bm{\Sigma}}_{1,n}^{(t)} \check{\bm{w}}_1^{(t)}x_n \in \mathbb{R}_{+}^{2H}, 	
\end{equation}
where $\check{\bm{\Sigma}}_{k,n}^{(t)}=\textnormal{diag}\big(\mathbb{I}\big\{\check{\bm{W}}_k^{(t)}\big(\prod_{l=1}^{k-2}  \check{\bm{\Sigma}}_{k-l,n}^{(t)}\check{\bm{W}}_{k-l}^{(t)}\big)\check{\bm{\Sigma}}_{1,n}^{(t)}\check{\bm{w}}_1^{(t)}x_n \\ \geq 0\big\}\big)\in\{0,1\}^{2H\times 2H}$ and $\check{\bm{\Sigma}}_{1,n}^{(t)}=\textnormal{diag}\big(\mathbb{I}\{\check{\bm{w}}_1^{(t)}x_n\geq 0\}\big)\in\{0,1\}^{2H\times 2H}$. Minimizing the training loss of this deep FNN and the mentioned deep ReLU FNN follows via (\ref{y_n_K_simplified_with_tanh}) and (\ref{y_n_K_simplified}), respectively, on a mini-batch-by-mini-batch basis.

After forward-propagating the $t$-th mini-batch $\mathcal{B}^{(t)}$ sampled randomly from $\mathcal{D}$, the training loss exhibited by the aforementioned deep ReLU FNN and deep FNN -- trained using hinge loss $\ell(t,y) \eqdef \max(0, 1-ty)$ -- is defined as    
\begin{equation}
	\label{prediction_Loss_defn}
	L_{\mathcal{B}^{(t)}}\big(\bm{W}^{(t)}\big) \eqdef  B^{-1} \sum_{i\in \mathcal{B}^{(t)}}  \ell\big(y_i,f_{\bm{W}^{(t)}}(x_i)\big),
\end{equation}
where $B\eqdef|\mathcal{B}^{(t)}|$, $\bm{W}^{(t)}$ is the concatenated weights of all layers, and $f_{\bm{W}^{(t)}}(x_n) = \bm{\Phi}^{(t)}(x_n)$ per (\ref{y_n_K_simplified}) or $f_{\bm{W}^{(t)}}(x_n) = \check{\bm{\Phi}}^{(t)}(x_n)$ per (\ref{y_n_K_simplified_with_tanh}). The deep FNNs' overall training error over $\mathcal{D}$ can be expressed as \cite{Wang_Learning_ReLU_FNN'2019}   
\begin{equation}
	\label{overall_loss_FNN_and_ReLU_FNN}
	R_{\mathcal{D}}\big(\bm{W}^{(t)}\big) \eqdef  N^{-1} \sum_{n=1}^N \mathbb{I}\{ y_n \neq \textnormal{sgn}(f_{\bm{W}^{(t)}}(x_n)) \}.
\end{equation}
To minimize (\ref{overall_loss_FNN_and_ReLU_FNN}), training of DL-based binary classifiers is conducted using the back-propagation algorithm to solve $\bm{W}^{*}\eqdef \sum_{t\in\mathbb{N}, \hspace{1mm} \mathcal{B}^{(t)} \subset \mathcal{D}}\argmin_{\bm{W}^{(t)}} L_{\mathcal{B}^{(t)}}\big(\bm{W}^{(t)}\big)$ while learning deep models that generalize better. To reveal the fundamental limits of such DL-based binary classification, we move on to the following problem formulation.
	
\section{Problem Formulation}
\label{sec: testing_performance_deep_FNN-Based_classifiers}
\subsection{Problems for ReLU FNNs-Based Binary Classifiers}
\label{subsec: Prob_ReLU_FNNs}
Let $\bm{\Phi}^{(T)}\eqdef\big[ [\bm{w}_1^{(T)}, \bm{0}], [\bm{W}_2^{(T)}, \bm{0}],  \ldots,[\bm{W}_{K-1}^{(T)}, \bm{0}], [\bm{w}_K^{(T)}, \bm{0}] \big]$ be a deep ReLU FNN trained using hinge loss and \texttt{SGD}, \texttt{SGD with momentum}, \texttt{RMSProp}, or \texttt{Adam} after being initialized by a Gaussian initializer such as \texttt{LeCun normal} or \texttt{He normal} \cite[Definition 10]{arXiv_Getu_Fundamental_Limits'23}. Note that $\bm{w}_{1}^{(T)}= \bm{w}_{1}^{(0)}+\sum_{t=0}^{T-1} \Delta\bm{w}_{1}^{(t)} \in\mathbb{R}^{2H}$, $\bm{W}_{k}^{(T)}= \bm{W}_{k}^{(0)}+\sum_{t=0}^{T-1} \Delta\bm{W}_{k}^{(t)}$, and
\begin{equation}
\label{the_K-th_weights_at_convergence}
\bm{w}_{K}^{(T)}= \bm{w}_{K}^{(0)}+\sum_{t=0}^{T-1} \Delta\bm{w}_{K}^{(t)} \in \mathbb{R}^{1\times 2H},
\end{equation}
where $T\in\mathbb{N}$; $\bm{w}_{1}^{(0)}$, $\bm{W}_{k}^{(0)}$, and $\bm{w}_{K}^{(0)}$ are weight matrices/vectors initialized with Gaussian initializers; $\Delta\bm{w}_{1}^{(t)}$, $\Delta\bm{W}_{k}^{(t)}$, and $\Delta\bm{w}_{K}^{(t)}$ are the $t$-th back-propagated error matrices/vectors. As to a collected weight matrix $\bm{W}^{(T)} \eqdef \big[\bm{w}_1^{(T)} \hspace{2mm} \bm{W}_2^{(T)} \ldots \bm{W}_{K-1}^{(T)} \hspace{2mm} (\bm{w}_K^{(T)})^\top \big]$ of all the layers' trained weights, the testing error over the testing set $\mathcal{T}\eqdef \big\{ (x_n, y_n) \big\}_{n=N+1}^{N+\tilde{N}}$ -- given that $x_n$ is the $n$-th input generated via (\ref{x_i__binary_class_problem}) and $y_n\in\{-1, 1\}$ is the $n$-th label for all\linebreak $n\in\{N+1, \ldots, N+\tilde{N}\}$ -- can be defined as    
\begin{equation}
	\label{testing_error_defn_1}
	R_{\mathcal{T}}\big(\bm{W}^{(T)}\big) \eqdef \tilde{N}^{-1} \sum_{n=N+1}^{N+\tilde{N}} \mathbb{I}\big\{ y_n \neq  \textnormal{sgn}\big(\bm{\Phi}^{(T)}(x_n)\big)   \big\}, 
\end{equation}
where $\tilde{N}\eqdef|\mathcal{T}|$. W.r.t. the right-hand side (RHS) of (\ref{testing_error_defn_1}), the underneath lemma ensues.  
\begin{lemma}
	\label{error_equivalence}
	For $y_n\in\{-1, 1\}$ and $n\in\mathbb{N}$,   
	\begin{equation}
		\label{equiv_eqn_1}
		\mathbb{I}\big\{ y_n \neq  \textnormal{sgn}\big(\bm{\Phi}^{(T)}(x_n)\big)   \big\} \equiv \mathbb{I}\big\{ y_n \bm{\Phi}^{(T)}(x_n) \leq 0  \big\}. 
	\end{equation}
	\proof We provide the proof in \cite[Appendix B, p. 53]{arXiv_Getu_Fundamental_Limits'23}. 
\end{lemma}

Substituting (\ref{equiv_eqn_1}) in the RHS of (\ref{testing_error_defn_1}) gives  
\begin{equation}
	\label{testing_error_defn}
	R_{\mathcal{T}}\big(\bm{W}^{(T)}\big)= \tilde{N}^{-1}\sum_{n=N+1}^{N+\tilde{N}} \mathbb{I}\big\{ y_n \bm{\Phi}^{(T)}(x_n) \leq 0  \big\}. 
\end{equation}
Hence, the average probability of misclassification error $\bar{P}_e$ manifested by $\bm{\Phi}^{(T)}$ becomes
\begin{equation}
	\label{P_e_average_defn}
	\bar{P}_e=\tilde{N}^{-1}\sum_{n=N+1}^{N+\tilde{N}} \mathbb{P}\big( y_n \bm{\Phi}^{(T)}(x_n) \leq 0  \big). 
\end{equation}
As for (\ref{P_e_average_defn}), the probability of point misclassification error $P_e$ manifested by $\bm{\Phi}^{(T)}$ equates to
\begin{equation}
\label{P_e_FNN_1}
P_e=\mathbb{P}\big( y_n \bm{\Phi}^{(T)}(x_n) \leq 0   \big). 
\end{equation}
The RHS of (\ref{P_e_FNN_1}) is simplified -- as done in \cite[eq. (39b)]{arXiv_Getu_Fundamental_Limits'23} to \cite[(40d)]{arXiv_Getu_Fundamental_Limits'23} -- using the \textit{law of total probability} \cite[Law 1]{arXiv_Getu_Fundamental_Limits'23} for equally likely labels that $\mathbb{P}(y_n=\pm1)=1/2$:
\begin{equation}
	\label{P_e_FNN_5}
	\hspace{-3mm}P_e=\frac{1}{2} \big[ \mathbb{P}\big( \bm{\Phi}^{(T)}(x_n) > 0 |\mathcal{H}_{-1} \big) + \mathbb{P}\big( \bm{\Phi}^{(T)}(x_n) < 0 |\mathcal{H}_1 \big) \big].	
\end{equation}
We formulate the following problems w.r.t. $\bm{y}_{K-1,n}^{(T)}$ defined via (\ref{y_n_K-1_simplified}) and $P_e$ given by (\ref{P_e_FNN_5}).  
\begin{problem}
	\label{Prob_perf_limts_ReLU_FNN_1}
	Characterize $\displaystyle\lim_{ (\| \bm{y}_{K-1,n}^{(T)} \||\mathcal{H}_{-1}, \| \bm{y}_{K-1,n}^{(T)} \| |\mathcal{H}_1 ) \to (\infty, \infty)} P_e$.
\end{problem}
\begin{problem}
	\label{Prob_perf_limts_ReLU_FNN_2}
	Quantify $\displaystyle\lim_{ (\| \bm{y}_{K-1,n}^{(T)} \||\mathcal{H}_{-1}, \| \bm{y}_{K-1,n}^{(T)} \||\mathcal{H}_1 ) \to (0, \infty)} P_e$.
\end{problem}

Problems \ref{Prob_perf_limts_ReLU_FNN_1}\&\ref{Prob_perf_limts_ReLU_FNN_2} are novel in the sense that a problem solver should address the interpretability of a trained deep ReLU FNN model while also circumventing the random gradients (with unknown distribution) that were back-propagated during training. 

\subsection{Problems for Binary Classifiers based on FNNs with ReLU and Tanh Activation}
\label{subsec: Prob_FNNs}
Let $\check{\bm{\Phi}}^{(T)} \eqdef \big[ [\check{\bm{w}}_1^{(T)}, \bm{0}], [\check{\bm{W}}_2^{(T)},  \bm{0}],  \ldots,[\check{\bm{W}}_{K-1}^{(T)}, \bm{0}], [\check{\bm{w}}_K^{(T)}, \bm{0}] \big]$ be a deep FNN trained using hinge loss and \texttt{SGD}, \texttt{SGD with momentum}, \texttt{RMSProp}, or \texttt{Adam} after being initialized with a Gaussian initializer such as \texttt{LeCun normal} or \texttt{He normal}. Note that $\check{\bm{w}}_{1}^{(T)}= \check{\bm{w}}_{1}^{(0)}+\sum_{t=0}^{T-1} \Delta\check{\bm{w}}_{1}^{(t)} \in\mathbb{R}^{2H}$, $\check{\bm{W}}_{k}^{(T)}= \check{\bm{W}}_{k}^{(0)}+\sum_{t=0}^{T-1} \Delta\check{\bm{W}}_{k}^{(t)} \in\mathbb{R}^{2H\times 2H}$, and
\begin{equation}
\label{the_K-th_weights_at_convergence_with_tanh}
\check{\bm{w}}_{K}^{(T)}= \check{\bm{w}}_{K}^{(0)}+\sum_{t=0}^{T-1} \Delta\check{\bm{w}}_{K}^{(t)} \in \mathbb{R}^{1\times 2H},
\end{equation}
where $\check{\bm{w}}_{1}^{(0)}$, $\check{\bm{W}}_{k}^{(0)}$, and $\check{\bm{w}}_{K}^{(0)}$ are weight matrices/vectors initialized with Gaussian initializers; $\Delta\check{\bm{w}}_{1}^{(t)}$, $\Delta\check{\bm{W}}_{k}^{(t)}$, and $\Delta\check{\bm{w}}_{K}^{(t)}$ are the $t$-th back-propagated error matrices/vectors.

Following (\ref{P_e_FNN_5}), the probability of point misclassification error $\check{P}_e$ manifested by $\check{\bm{\Phi}}^{(T)}$ is given by
\begin{equation}
	\label{P_e_FNN_5_with_tanh}
	\hspace{-3mm}\check{P}_e=\frac{1}{2} \big[ \mathbb{P}\big( \check{\bm{\Phi}}^{(T)}(x_n) > 0 |\mathcal{H}_{-1} \big) + \mathbb{P}\big( \check{\bm{\Phi}}^{(T)}(x_n) < 0 |\mathcal{H}_1 \big) \big].	
\end{equation} 
We formulate the following problems for (\ref{P_e_FNN_5_with_tanh}) w.r.t. $\check{\bm{y}}_{K-1,n}^{(T)}$, which is defined via (\ref{y_n_K-1_simplified_with_tanh}), and $\check{P}_e$ given by (\ref{P_e_FNN_5_with_tanh}).  
\begin{problem}
	\label{Prob_perf_limts_FNN_with_ReLU_and_Hyperbolic_Tangent_1}
	Determine $\displaystyle\lim_{ (\|\check{\bm{y}}_{K-1,n}^{(T)} \| |\mathcal{H}_{-1}, \| \check{\bm{y}}_{K-1,n}^{(T)} \| |\mathcal{H}_1 ) \to (\infty, \infty)} \check{P}_e$.
\end{problem}
\begin{problem}
	\label{Prob_perf_limts_FNN_with_ReLU_and_Hyperbolic_Tangent_2}
	Derive $\displaystyle\lim_{ (\| \check{\bm{y}}_{K-1,n}^{(T)} \| |\mathcal{H}_{-1}, \| \check{\bm{y}}_{K-1,n}^{(T)} \| |\mathcal{H}_1 ) \to (0, \infty)} \check{P}_e$.
\end{problem}
Problems \ref{Prob_perf_limts_FNN_with_ReLU_and_Hyperbolic_Tangent_1}\&\ref{Prob_perf_limts_FNN_with_ReLU_and_Hyperbolic_Tangent_2} are also novel in the sense that a problem solver must tackle the interpretability of a trained deep FNN model with ReLU and Tanh activation while also circumventing the random gradients (with unknown distribution) that were back-propagated during training. 

\section{Asymptotic Testing Performance Limits}
\label{sec: asymptotic_performance_limits_FNN_Based_binary_classifiers}
\subsection{Performance Limits of Binary Classifiers that are Based on Deep ReLU FNNs}
\label{subsec: asymptotic_performance_limits_of_ReLU_FNN_based_binary_classifiers}
The asymptotic testing performance limits of $\bm{\Phi}^{(T)}$ follow.   
\begin{theorem}
	\label{thm_performance_limits_ReLU_FNN_based_detector}
	Consider a (deep) ReLU FNN $\bm{\Phi}^{(T)}\eqdef\big[ [\bm{w}_1^{(T)}, \bm{0}], [\bm{W}_2^{(T)}, \bm{0}],  \ldots,[\bm{W}_{K-1}^{(T)}, \bm{0}],  [\bm{w}_K^{(T)}, 0] \big]$ -- such that $\bm{w}_K^{(T)}\in\mathbb{R}^{1\times 2H}$, $\bm{W}_{K-k}^{(T)}\in\mathbb{R}^{2H\times 2H}$ for all $k\in[K-2]$, and $\bm{w}_1^{(T)}\in\mathbb{R}^{2H}$ -- trained over $\mathcal{D}\eqdef \big\{ (x_n, y_n) \big\}_{n=1}^{N}$ using the settings of Sec. \ref{subsec: Prob_ReLU_FNNs}, and to be tested over $\mathcal{T}\eqdef \big\{ (x_n, y_n) \big\}_{n=N+1}^{N+\tilde{N}}$, given $H, K, N, \tilde{N}, T\in\mathbb{N}$. For $N, \tilde{N}, T, K\geq 2$, and $H<\infty$: 
	\begin{subequations}
		\begin{align}
			\label{inference_1_deep_ReLU_FNN}
			\lim_{\| \bm{y}_{K-1,n}^{(T)}\| \to \infty} P_e &\leq 1/2  \\
			\label{inference_2_deep_ReLU_FNN}
			 \lim_{( \| \bm{y}_{K-1,n}^{(T)}\| | \mathcal{H}_{-1},  \| \bm{y}_{K-1,n}^{(T)}\| | \mathcal{H}_{1} ) \to (0, \infty)} P_e  & \leq 1/2, 
		\end{align}
	\end{subequations}
	where (\ref{inference_1_deep_ReLU_FNN}) and (\ref{inference_2_deep_ReLU_FNN}) are valid for all $n\in \{ N+1, \ldots, N+\tilde{N}\}$.
	
	\proof The proof is given in the supplementary material. 
\end{theorem}

\begin{remark}
\label{Rem_Thm_1_1}
Theorem \ref{thm_performance_limits_ReLU_FNN_based_detector} is valid for any training/testing dataset size, network depth, or (finite) network width, since (\ref{inference_1_deep_ReLU_FNN}) and (\ref{inference_2_deep_ReLU_FNN}) are valid for any $N, \tilde{N}, T, K\geq 2$, and $H<\infty$.
\end{remark}
\begin{remark}
\label{Rem_Thm_1_2}
Theorem \ref{thm_performance_limits_ReLU_FNN_based_detector} applies to any binary classification AI/ML problem solved using a (deep) ReLU FNN that has been trained with hinge loss and \texttt{SGD}, \texttt{SGD with momentum}, \texttt{RMSProp}, or \texttt{Adam} after being initialized with a Gaussian initializer such as \texttt{LeCun normal} or \texttt{He normal}, because it does not make any assumptions w.r.t. the input hypotheses.
\end{remark}
\begin{remark}
\label{Rem_Thm_1_3}
If $\|\bm{y}_{K-1,n}^{(T)}\| \to \infty$ or $( \| \bm{y}_{K-1,n}^{(T)}\| | \mathcal{H}_{-1},  \linebreak \| \bm{y}_{K-1,n}^{(T)}\| | \mathcal{H}_{1} ) \to (0, \infty)$, the exhibited probability of misclassification error will be upper bounded by 1/2, where the binary classifier would be in a coin-tossing mode.
\end{remark}
  
\subsection{Performance Limits of Binary Classifiers based on FNNs with ReLU and Tanh Activation}
\label{subsec: asymptotic_performance_limits_FNN_based_binary_classifiers_with_ReLU_and_tanh}
The asymptotic testing performance limits of $\check{\bm{\Phi}}^{(T)}$ ensue. 
\begin{theorem}
	\label{thm_performance_limits_FNN_based_detector_with_tanh}
	Consider a (deep) FNN with ReLU and Tanh activation $\check{\bm{\Phi}}^{(T)}\eqdef\big[ [\check{\bm{w}}_1^{(T)}, \bm{0}], [\check{\bm{W}}_2^{(T)}, \bm{0}],  \ldots,[\check{\bm{W}}_{K-1}^{(T)}, \bm{0}],  [\check{\bm{w}}_K^{(T)}, 0] \big]$ -- such that $\check{\bm{w}}_K^{(T)}\in\mathbb{R}^{1\times 2H}$, $\check{\bm{W}}_{K-k}^{(T)}\in\mathbb{R}^{2H\times 2H}$ $\forall k\in[K-2]$, and $\check{\bm{w}}_1^{(T)}\in\mathbb{R}^{2H}$ -- trained over $\mathcal{D}\eqdef \big\{ (x_n, y_n) \big\}_{n=1}^{N}$ per the settings of Sec. \ref{subsec: Prob_FNNs}, and to be tested over $\mathcal{T}\eqdef \big\{ (x_n, y_n) \big\}_{n=N+1}^{N+\tilde{N}}$, given $H, K, N, \tilde{N}, T\in\mathbb{N}$. For $N, \tilde{N}, T, K\geq 2$ and $H<\infty$: 
	\begin{subequations}
		\begin{align}
			\label{inference_1_deep_FNN_with_tanh}
			\lim_{\| \check{\bm{y}}_{K-1,n}^{(T)} \|  \to \infty} \check{P}_e  & \leq 1/2   \\
			\label{inference_2_deep_FNN_with_tanh}
			\lim_{ (\| \check{\bm{y}}_{K-1,n}^{(T)} \| |\mathcal{H}_{-1},  \| \check{\bm{y}}_{K-1,n}^{(T)} \| |\mathcal{H}_{1}) \to (0, \infty)} \check{P}_e& \leq 1/2,
		\end{align}
	\end{subequations}
where (\ref{inference_1_deep_FNN_with_tanh}) and (\ref{inference_2_deep_FNN_with_tanh}) are true for all $n\in \{ N+1,  \ldots, N+\tilde{N}\}$.
	\proof The proof is provided in the supplementary material.    
\end{theorem}

\begin{remark}
\label{Rem_Thm_2_1}
Theorem \ref{thm_performance_limits_FNN_based_detector_with_tanh} is valid for any training/testing dataset size, network depth, or (finite) network width, because (\ref{inference_1_deep_FNN_with_tanh}) and (\ref{inference_2_deep_FNN_with_tanh}) are valid for any $N, \tilde{N}, T, K\geq 2$, and $H<\infty$.
\end{remark}

\begin{remark}
\label{Rem_Thm_2_2}
Theorem \ref{thm_performance_limits_FNN_based_detector_with_tanh} is applicable to any binary classification AI/ML problem solved using a deep FNN with ReLU and Tanh activation that has been trained with hinge loss and \texttt{SGD}, \texttt{SGD with momentum}, \texttt{RMSProp}, or \texttt{Adam} after being initialized with a Gaussian initializer such as \texttt{LeCun normal} or \texttt{He normal}, since it does not make any assumptions w.r.t. the input hypotheses.
\end{remark}

\begin{remark}
\label{Rem_Thm_2_3}
If $\|\check{\bm{y}}_{K-1,n}^{(T)}\| \to \infty$ or $( \| \check{\bm{y}}_{K-1,n}^{(T)}\| | \mathcal{H}_{-1}, \linebreak \| \check{\bm{y}}_{K-1,n}^{(T)}\| | \mathcal{H}_{1} ) \to (0, \infty)$, the manifested probability of misclassification error will be upper bounded by 1/2, where the binary classifier would be in a coin-tossing mode.
\end{remark}

\section{Computer Experiments}
\label{sec: results}
\subsection{DL-Based Binary Classification Settings}
\label{subsec: experimental_context}
In view of the generic binary hypothesis testing of (\ref{x_i__binary_class_problem}), we opt for a classical optimum detection problem involving a binary phase shift keying (BPSK) signal received over an AWGN channel. For a received AWGN-contaminated BPSK signal having a bit duration of $T_s$, we consider the received passband signal is down-converted to the received baseband signal, denoted by $r(t)$, which is sampled per the Nyquist rate $f_s=1/T_s$ to produce the sampled baseband signal $r[n]\eqdef r(nT_s)$ given for $n\in\mathbb{N}$ by            
\begin{equation}
	\label{r_n_BPSK}
	r[n]=\begin{cases}
		-\sqrt{\mathcal{E}_b} + z[n] &: \mathcal{H}_{-1} \\
		\sqrt{\mathcal{E}_b} + z[n] & : \mathcal{H}_1,  
	\end{cases}
\end{equation}
where $\mathcal{E}_b$ is the transmitted energy per bit; $s_0=-\sqrt{\mathcal{E}_b}$ and $s_1=\sqrt{\mathcal{E}_b}$ are the equally likely samples of the transmitted BPSK signal pertaining to bit 0 and bit 1, respectively; $z[n]\sim\mathcal{N}(0,N_0/2)$ denotes the AWGN samples obtained from the AWGN process $z(t)$ as $z[n] \eqdef z(nT_s)$, where $N_0$ is the power spectral density of the noise; and $\mathcal{H}_{-1}$ and $\mathcal{H}_1$ are hypotheses corresponding to the transmission of $s_0$ and $s_1$, respectively. As to the problem in (\ref{r_n_BPSK}), an optimum BPSK detector's manifested probability of bit error ($\tilde{P}_e$) is given by $\tilde{P}_e=\mathcal{Q}(\sqrt{2\gamma_b})$, as we proved in \cite[Appendix E, p. 65-68]{arXiv_Getu_Fundamental_Limits'23}, where $\gamma_b$ is the signal-to-noise ratio (SNR) per bit. This was used as a baseline for our computer experiments assessing the performance of DL-based BPSK detectors.   
\begin{table*}[!htb]
	\centering
	\begin{tabular}{ | l | c | c | c | c | c | c | c | }
		\hline
		\hspace{1mm}SNR & \multicolumn{1}{ c | }{Optimal} & \multicolumn{3}{ c | }{ReLU FNN-based BPSK detector}  & \multicolumn{3}{ c | }{ReLU FNN-based BPSK detector}  \\
		$[$in dB$]$ & \multicolumn{1}{ c | }{detector} & \multicolumn{3}{ c | }{$(K, H)=(3,3)$}  & \multicolumn{3}{ c | }{$(K, H)=(3,9)$}  \\
		\hline
		$\gamma_b$ $[$dB$]$ & $\mathcal{Q}(\sqrt{2\gamma_b})$ & All-SNR-T & High-SNR-T & Low-SNR-T & All-SNR-T & High-SNR-T & Low-SNR-T    \\  \hline  
		0 & 0.0786 & 0.56448001 & 0.75902 & 0.56411999 & 0.56411999 & 0.59154001 & 0.56402001      \\  \hline
		5 & 0.0060 & 0.50569999 & 0.57170001 & 0.50566 & 0.50566 & 0.51100001 & 0.50564      \\  \hline
		10 & $3.8721\times 10^{-6}$ & 0.50075999 & 0.50128001 & 0.50075999 & 0.50075999 & 0.50075999 &  0.50075999    \\  \hline
		15 & $9.1240\times 10^{-16}$ & 0.50338  & 0.50338 & 0.50338 & 0.50338  & 0.50338  &  0.50338     \\  \hline
		20 & $1.0442 \times 10^{-45} $ & 0.49835998 & 0.49835998 & 0.49835998 & 0.49835998 & 0.49835998 &  0.49835998    \\  \hline
		25 & $7.3070 \times 10^{-140} $ & 0.49908 & 0.49908 & 0.49908 & 0.49908 & 0.49908 &  0.49908    \\  \hline
		30 & 0 & 0.49677998 & 0.49677998 & 0.49677998 & 0.49677998 & 0.49677998 &  0.49677998    \\  \hline
		35 & 0 & 0.50049999 & 0.50049999 & 0.50049999 & 0.50049999 & 0.50049999 &  0.50049999    \\  \hline
	\end{tabular}   \\ [4mm]
	\caption{Numerical results I on bit error rate manifested by optimal BPSK detector and ReLU FNN-based BPSK detectors.}
	\label{computer_experiment_results_I}
\end{table*}
\begin{table*}[!t]
	\centering
	\begin{tabular}{ | l | c | c | c | c | c | c | c | }
		\hline
		\hspace{1mm}SNR & \multicolumn{1}{ c | }{Optimal} & \multicolumn{3}{ c | }{ReLU FNN-based BPSK detector}  & \multicolumn{3}{ c | }{ReLU FNN-based BPSK detector}  \\
		$[$in dB$]$ & \multicolumn{1}{ c | }{detector} & \multicolumn{3}{ c | }{$(K, H)=(7,7)$}  & \multicolumn{3}{ c | }{$(K, H)=(7,14)$}  \\
		\hline
		$\gamma_b$ $[$dB$]$ & $\mathcal{Q}(\sqrt{2\gamma_b})$ & All-SNR-T & High-SNR-T & Low-SNR-T & All-SNR-T & High-SNR-T & Low-SNR-T    \\  \hline  
		0 & 0.0786 & 0.56362 &0.86848 &0.56345999 &0.56378001 &0.55552 & 0.56332001     \\  \hline
		5 & 0.0060 & 0.50562 &0.65803999 &0.50562 &0.50564 &0.50426 &0.50558001      \\  \hline
		10 & $3.8721\times 10^{-6}$ & 0.50075999 &0.50454 &0.50075999 &0.50075999 &0.50075999 &0.50075999      \\  \hline
		15 & $9.1240\times 10^{-16}$ & 0.50338 &0.50338 &0.50338 &0.50338 &0.50338 &0.50338      \\  \hline
		20 & $1.0442 \times 10^{-45} $ &0.49835998 &0.49835998 &0.49835998 &0.49835998 &0.49835998 &0.49835998      \\  \hline
		25 & $7.3070 \times 10^{-140} $ &0.49908 &0.49908 &0.49908 &0.49908 &0.49908 &0.49908      \\  \hline
		30 & 0 &0.49677998 &0.49677998 &0.49677998 &0.49677998 &0.49677998 &0.49677998      \\  \hline
		35 & 0 &0.50049999 &0.50049999 &0.50049999 &0.50049999 &0.50049999 &0.50049999      \\  \hline
	\end{tabular}   \\ [4mm]
	\caption{Numerical results II on bit error rate exhibited by optimal BPSK detector and ReLU FNN-based BPSK detectors.}
	\label{computer_experiment_results_II}
\end{table*}
\begin{table*}[!t]
	\centering
	\begin{tabular}{ | l | c | c | c | c | c | c | c | }
		\hline
		\hspace{1mm}SNR & \multicolumn{1}{ c | }{Optimal} & \multicolumn{3}{ c | }{ReLU FNN-based BPSK detector}  & \multicolumn{3}{ c | }{ReLU FNN-based BPSK detector}  \\
		$[$in dB$]$ & \multicolumn{1}{ c | }{detector} & \multicolumn{3}{ c | }{$(K, H)=(9,9)$}  & \multicolumn{3}{ c | }{$(K, H)=(9, 18)$}  \\
		\hline
		$\gamma_b$ $[$dB$]$ & $\mathcal{Q}(\sqrt{2\gamma_b})$ & All-SNR-T & High-SNR-T & Low-SNR-T & All-SNR-T & High-SNR-T & Low-SNR-T    \\  \hline  
		0 & 0.0786 &0.56356001 &0.64416 &0.56376001 &0.56417999 &0.55226001 &0.565      \\  \hline
		5 & 0.0060 &0.50562 &0.52410001 &0.50564 &0.50567999 &0.50376001 &0.50588      \\  \hline
		10 & $3.8721\times 10^{-6}$ &0.50075999 &0.50080001 &0.50075999 &0.50075999 &0.50075999 &0.50075999      \\  \hline
		15 & $9.1240\times 10^{-16}$ &0.50338 &0.50338 &0.50338 &0.50338 &0.50338 &0.50338      \\  \hline
		20 & $1.0442 \times 10^{-45} $ &0.49835998 &0.49835998 &0.49835998 &0.49835998 &0.49835998 &0.49835998      \\  \hline
		25 & $7.3070 \times 10^{-140} $ &0.49908 &0.49908 &0.49908 &0.49908 &0.49908 &0.49908      \\  \hline
		30 & 0 &0.49677998 &0.49677998 &0.49677998 &0.49677998 &0.49677998 &0.49677998      \\  \hline
		35 & 0 &0.50049999 &0.50049999 &0.50049999 &0.50049999 &0.50049999 &0.50049999      \\  \hline
	\end{tabular}   \\ [4mm]
	\caption{Numerical results III on bit error rate manifested by optimal BPSK detector and ReLU FNN-based BPSK detectors.}
	\label{computer_experiment_results_III}
\end{table*}

\subsection{Materials and Methods}
\label{subsec: materials_and_methods}
After using \textsc{MATLAB}\textsuperscript{\textregistered} to generate our training and testing sets, our computer experiments were carried out on the NIST GPU cluster named \textit{Enki} using Keras with TensorFlow as a backend. Our training sets $\mathcal{D}_1$ and $\mathcal{D}_2$ were generated by combining the training data generated for $\gamma_b\in\{0, 5, 10, 15, 20, 25, 30, 35\}$ dB. For $\gamma_b^i=5(i-1)$ dB and $i\in[8]$, we generated training sets $\mathcal{D}_1^i$ and $\mathcal{D}_2^i$ made of $2\times 10^5$ inputs equal to the received baseband signal samples produced at $\gamma_b^i$ and equally likely labels chosen from $\{-1,-1\}$ as $\mathcal{D}_1^i\eqdef \big\{ (x_n^i, y_n^i) \big\}_{n=1}^{200,000}$ and $\mathcal{D}_2^i\eqdef \big\{ (x_{\tilde{n}}^i, y_{\tilde{n}}^i) \big\}_{\tilde{n}=1}^{200,000}$ such that $x_n^i \eqdef r[n] |_{\gamma_b=\gamma_b^i}$ for $r[n]$ defined in (\ref{r_n_BPSK}), $y_n^i\in\{-1,1\}$, $x_{\tilde{n}}^i \eqdef r[\tilde{n}] |_{\gamma_b=\gamma_b^i}$ for $r[\tilde{n}]$ defined via (\ref{r_n_BPSK}), and $y_{\tilde{n}}^i\in\{-1,1\}$. For the generation of $x_{\tilde{n}}^i \eqdef r[\tilde{n}] |_{\gamma_b=\gamma_b^i}$ and $x_n^i \eqdef r[n] |_{\gamma_b=\gamma_b^i}$ w.r.t. $\gamma_b^i=\mathcal{E}_b^i/N_0$, we set $N_0=1$ W/Hz such that $\gamma_b^i=\mathcal{E}_b^i$, $z[n]\sim\mathcal{N}(0,1/2)$, and $z[\tilde{n}]\sim\mathcal{N}(0,1/2)$. We used these settings to generate $\mathcal{D}_1^i$ and $\mathcal{D}_2^i$ -- for all $i\in[8]$ -- and formed $\mathcal{D}_1$ and $\mathcal{D}_2$ as $\mathcal{D}_1\eqdef \big\{ \mathcal{D}_1^1 \cup \mathcal{D}_1^2 \cup \cdots \cup  \mathcal{D}_1^7 \cup \mathcal{D}_1^8  \big\}$ and $\mathcal{D}_2\eqdef \big\{ \mathcal{D}_2^1 \cup \mathcal{D}_2^2 \cup \cdots \cup  \mathcal{D}_2^7 \cup \mathcal{D}_2^8  \big\}$, respectively. Similarly, we generated a testing set $\mathcal{T}$ as $\mathcal{T}\eqdef\big\{ \mathcal{T}^1 \cup \mathcal{T}^2 \cup \cdots \cup  \mathcal{T}^7 \cup \mathcal{T}^8  \big\}$ for $\mathcal{T}^i$, $i\in[8]$, being the testing set generated at SNR $\gamma_b^i$ such that $\mathcal{T}^i\eqdef \big\{ (x_n^i, y_n^i) \big\}_{n=200,001}^{250,000}$ for $x_n^i \eqdef r[n] |_{\gamma_b=\gamma_b^i}$, $y_n^i\in\{-1,1\}$, and $\gamma_b^i=5(i-1)$ dB. We then uploaded $\mathcal{T}$, $\mathcal{D}_1$, and $\mathcal{D}_2$ into Enki and implemented three training strategies, namely all SNR training (All-SNR-T), high SNR training (High-SNR-T), and low SNR training (Low-SNR-T).   

For All-SNR-T, we formed a training and validation set $\mathcal{D}$ as $\mathcal{D}=\mathcal{D}_1$. For High-SNR-T, we generated a training and validation set $\mathcal{D}_{\textnormal{high}}$ as $\mathcal{D}_{\textnormal{high}}=\mathcal{D}_{1, \textnormal{high}} \cup \mathcal{D}_{2, \textnormal{high}}$, where $\mathcal{D}_{1, \textnormal{high}} \eqdef \big\{ \mathcal{D}_1^5 \cup \mathcal{D}_1^6 \cup \mathcal{D}_1^7 \cup \mathcal{D}_1^8  \big\}$, $\mathcal{D}_{2, \textnormal{high}} \eqdef \big\{ \mathcal{D}_2^5 \cup \mathcal{D}_2^6 \cup \mathcal{D}_2^7 \cup \mathcal{D}_2^8  \big\}$, and $|\mathcal{D}_{1, \textnormal{high}}|=|\mathcal{D}|$. For Low-SNR-T, we created a training and validation set $\mathcal{D}_{\textnormal{low}}$ as $\mathcal{D}_{\textnormal{low}}=\mathcal{D}_{1, \textnormal{low}} \cup \mathcal{D}_{2, \textnormal{low}}$, where $\mathcal{D}_{1, \textnormal{low}} \eqdef \big\{ \mathcal{D}_1^1 \cup \mathcal{D}_1^2 \cup \mathcal{D}_1^3 \cup \mathcal{D}_1^4  \big\}$, $\mathcal{D}_{2, \textnormal{low}} \eqdef \big\{ \mathcal{D}_2^1 \cup \mathcal{D}_2^2 \cup \mathcal{D}_2^3 \cup \mathcal{D}_2^4  \big\}$, and $|\mathcal{D}_{1, \textnormal{low}}|=|\mathcal{D}|$. Using $\mathcal{D}_{\textnormal{low}}$, $\mathcal{D}_{\textnormal{high}}$, and $\mathcal{D}$, we executed exhaustive training that led to the optimized (hyper)parameters of \cite[Tables VI\&VII]{arXiv_Getu_Fundamental_Limits'23} for the best performing ReLU FNN models and FNN models with ReLU and Tanh activation, respectively. Besides, we used four \texttt{KERAS callbacks} : \texttt{TensorBoard}, \texttt{ModelCheckpoint}, \texttt{EarlyStopping}, and \texttt{ReduceLROnPlateau} \cite[Ch. 7]{Chollet_DL_with_Python'18}. \texttt{EarlyStopping} and \texttt{ReduceLROnPlateau} were set to have patience over 20 and 40 epochs, respectively. \texttt{ReduceLROnPlateau} was set to reduce the learning rate by 0.5 for every 20 epochs that led to performance stagnation. Finally, we tested the BPSK detection performance of trained (deep) ReLU FNNs and (deep) FNNs with ReLU and Tanh activation by varying $(K, H)$ for each testing SNR, leading to the results reported in Sections \ref{subsec: main_results_ReLU_FNNs} and \ref{subsec: main_results_FNNs}.        

\subsection{Results for ReLU FNNs}
\label{subsec: main_results_ReLU_FNNs}
Using the (hyper)parameters of \cite[Table VI]{arXiv_Getu_Fundamental_Limits'23}, we produced the ReLU FNN-based BPSK detectors' testing and theory validation results that are presented in Sections \ref{subsubsec: testing_results_ReLU_FNNs} and \ref{subsubsec: theory_validation_results_ReLU_FNNs}, respectively. 
\subsubsection{Testing Results}
\label{subsubsec: testing_results_ReLU_FNNs}
Tables \ref{computer_experiment_results_I}-\ref{computer_experiment_results_III} contrast the detection performance of the optimum BPSK detector and ReLU FNN-based BPSK detectors. Table \ref{computer_experiment_results_I} confirms that shallow ReLU FNNs with $K=3$ perform relatively well with Low-SNR-T better than with All-SNR-T or High-SNR-T for $\gamma_b\in\{0,5\}$ dB. Tables \ref{computer_experiment_results_II} and \ref{computer_experiment_results_III} validate that deep ReLU FNNs with $K\in\{7,9\}$ perform better with All-SNR-T than with Low-SNR-T or High-SNR-T for $\gamma_b\in\{0,5\}$ dB. Tables \ref{computer_experiment_results_I}-\ref{computer_experiment_results_III} substantiate that High-SNR-T\linebreak produces the worst performance for  $\gamma_b\in\{0,5\}$ dB; ReLU FNNs perform identically with All-SNR-T, High-SNR-T, and Low-SNR-T for $\gamma_b>5$ dB, irrespective of the networks' depth and width.       

The minimum probability of bit error exhibited by a ReLU FNN-based BPSK detector irrespective of its width, depth, and training scheme is almost 0.5 as attested by Tables \ref{computer_experiment_results_I}-\ref{computer_experiment_results_III}, which illustrate that a ReLU FNN-based BPSK detector trained with hinge loss and \texttt{Adam} fails to learn optimum BPSK detection. At high SNRs, a sampled BPSK signal is likely a large positive or negative number when bit 1 or bit 0 is transmitted over an AWGN channel, respectively.\linebreak This detection problem is thus a simple inference problem that can be solved by using the decision threshold $\tau_{\textnormal{th}}=0$. Nonetheless, our best-trained deep (or shallow) ReLU FNNs fail at this common-sense inference task. This may be because FNNs (and deep networks in general) are poor at solving (common-sense) inference tasks \cite{DL_Appraisal_normal_Marcus'18} such as BPSK signal detection at high SNRs.    
\begin{figure}[htb!]
	\centering
	\includegraphics[scale=0.45]{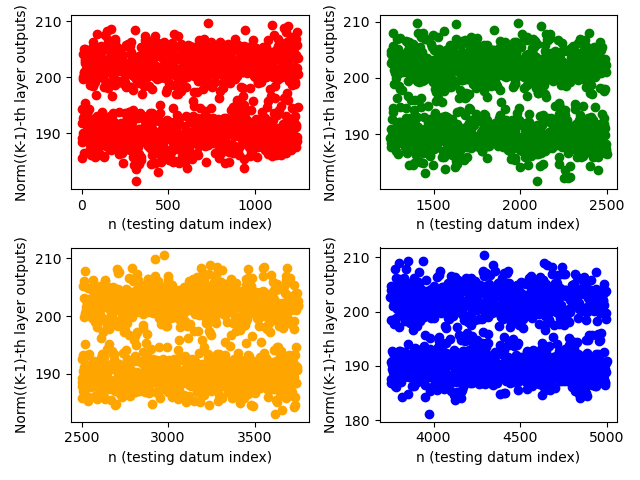}
	\caption{A scatter plot of $\| \bm{y}_{K-1,n}^{(T)} \|$ versus $n$ under All-SNR-T, $(K, H)=(8,8)$, and testing at 35 dB: the computed testing $P_e$ at 35 dB is $P_e=0.50049999$.}
	\label{fig: ReLU_FNN_all_SNRs_Norm_of_the_penultimate_layer_output_vectors_K_8_H_8}
\end{figure}
\begin{figure}[htb!]
	\centering
	\includegraphics[scale=0.45]{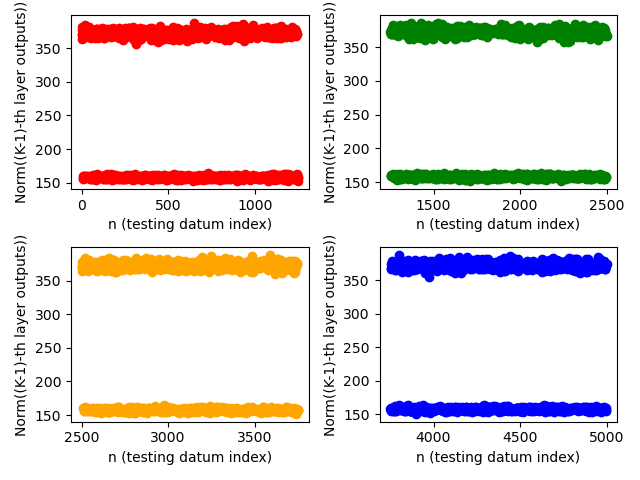}
	\caption{A scatter plot of $\| \bm{y}_{K-1,n}^{(T)} \|$ versus $n$ under High-SNR-T, $(K, H)=(8,8)$, and testing at 35 dB: the computed testing $P_e$ at 35 dB is $P_e=0.50049999$.}
	\label{fig: ReLU_FNN_high_SNR_Norm_of_the_penultimate_layer_output_vectors_K_8_H_8}
\end{figure} 
\begin{figure}[htb!]
	\centering
	\includegraphics[scale=0.45]{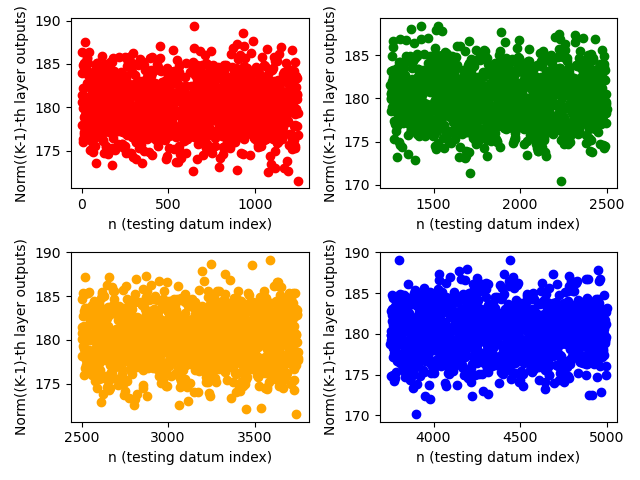}
	\caption{A scatter plot of $\| \bm{y}_{K-1,n}^{(T)} \|$ versus $n$ under Low-SNR-T, $(K, H)=(8,8)$, and testing at 35 dB: the computed testing $P_e$ at 35 dB is $P_e=0.50049999$.}
	\label{fig: ReLU_FNN_low_SNR_Norm_of_the_penultimate_layer_output_vectors_K_8_H_8}
\end{figure}

\subsubsection{Theory Validation Results}
\label{subsubsec: theory_validation_results_ReLU_FNNs}
For shallow ReLU FNN-based BPSK detectors with $(K, H)=(8,8)$, Figs. \ref{fig: ReLU_FNN_all_SNRs_Norm_of_the_penultimate_layer_output_vectors_K_8_H_8}-\ref{fig: ReLU_FNN_low_SNR_Norm_of_the_penultimate_layer_output_vectors_K_8_H_8} show the scatter plots of $\| \bm{y}_{K-1,n}^{(T)} \|$ versus $n$ subject to All-SNR-T, High-SNR-T, and Low-SNR-T, respectively. These scatter plots do not exactly corroborate Theorem \ref{thm_performance_limits_ReLU_FNN_based_detector} since $\| \bm{y}_{K-1,n}^{(T)} \|$ doesn't approach infinity or zero, though the testing $P_e$ equals 0.5 in all cases.
\begin{figure}[htb!]
	\centering
	\includegraphics[scale=0.45]{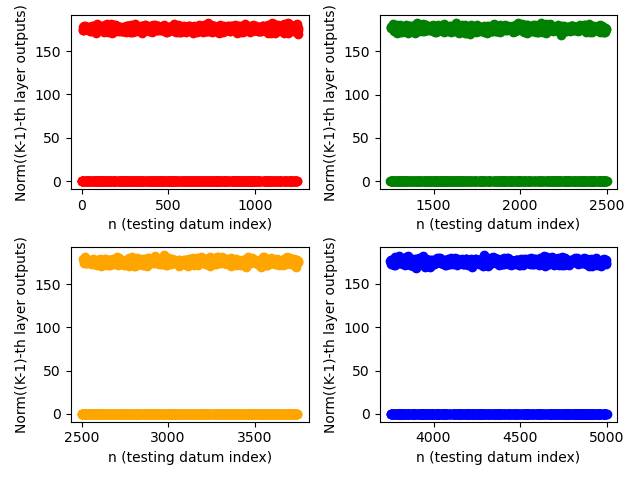}
	\caption{A scatter plot of $\| \bm{y}_{K-1,n}^{(T)} \|$ versus $n$ under All-SNR-T, $(K, H)=(16,16)$, and testing at 35 dB: the computed testing $P_e$ at 35 dB is $P_e=1.0$.}
	\label{fig: ReLU_FNN_all_SNRs_Norm_of_the_penultimate_layer_output_vectors_K_16_H_16}
\end{figure}

\begin{figure}[htb!]
	\centering
	\includegraphics[scale=0.45]{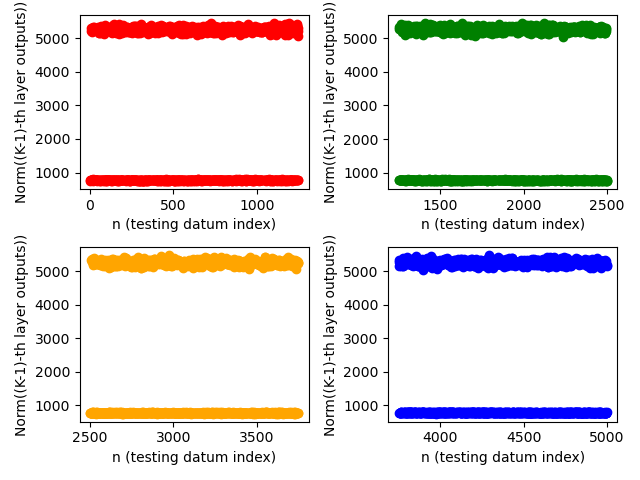}
	\caption{A scatter plot of $\| \bm{y}_{K-1,n}^{(T)} \|$ versus $n$ under High-SNR-T, $(K, H)=(16,16)$, and testing at 35 dB: the computed testing $P_e$ at 35 dB is $P_e=0.50049999$.}
	\label{fig: ReLU_FNN_high_SNR_Norm_of_the_penultimate_layer_output_vectors_K_16_H_16}
\end{figure} 

\begin{figure}[htb!]
	\centering
	\includegraphics[scale=0.45]{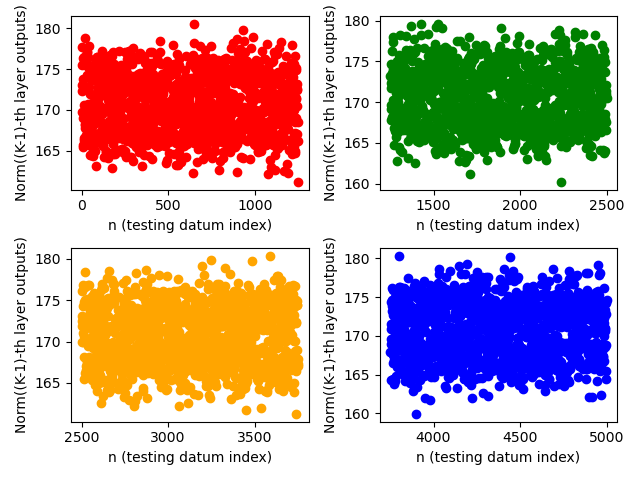}
	\caption{A scatter plot of $\| \bm{y}_{K-1,n}^{(T)} \|$ versus $n$ under Low-SNR-T, $(K, H)=(16,16)$, and testing at 35 dB: the computed testing $P_e$ at 35 dB is $P_e=0.50049999$.}
	\label{fig: ReLU_FNN_low_SNR_Norm_of_the_penultimate_layer_output_vectors_K_16_H_16}
\end{figure}   

As for deep ReLU FNN-based BPSK detectors with $(K, H)=(16,16)$, Figs. \ref{fig: ReLU_FNN_all_SNRs_Norm_of_the_penultimate_layer_output_vectors_K_16_H_16}-\ref{fig: ReLU_FNN_low_SNR_Norm_of_the_penultimate_layer_output_vectors_K_16_H_16} show the scatter plots of $\| \bm{y}_{K-1,n}^{(T)} \|$ versus $n$ subject to All-SNR-T, High-SNR-T, and Low-SNR-T, respectively. Fig. \ref{fig: ReLU_FNN_high_SNR_Norm_of_the_penultimate_layer_output_vectors_K_16_H_16} corroborates Theorem \ref{thm_performance_limits_ReLU_FNN_based_detector} since the obtained $P_e$ equals 0.5 and $\| \bm{y}_{K-1,n}^{(T)} \|$ approaches either zero or a very large number. However, Theorem \ref{thm_performance_limits_ReLU_FNN_based_detector} is not validated by Figs. \ref{fig: ReLU_FNN_all_SNRs_Norm_of_the_penultimate_layer_output_vectors_K_16_H_16}\&\ref{fig: ReLU_FNN_low_SNR_Norm_of_the_penultimate_layer_output_vectors_K_16_H_16}, which are among the results that cannot be interpreted by Theorem \ref{thm_performance_limits_ReLU_FNN_based_detector}.    

\begin{figure}[htb!]
	\centering
	\includegraphics[scale=0.45]{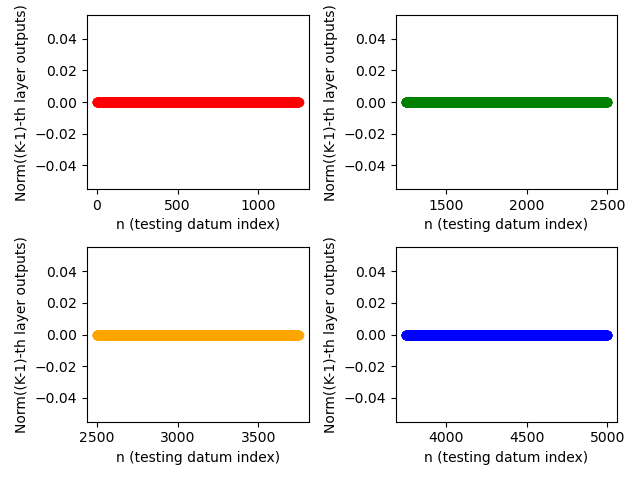}
	\caption{A scatter plot of $\| \bm{y}_{K-1,n}^{(T)} \|$ versus $n$ under All-SNR-T, $(K, H)=(20,100)$, and testing at 35 dB: the computed testing $P_e$ at 35 dB is $P_e=1.0$.}
	\label{fig: ReLU_FNN_all_SNRs_Norm_of_the_penultimate_layer_output_vectors_K_20_H_100}
\end{figure}

\begin{figure}[htb!]
	\centering
	\includegraphics[scale=0.45]{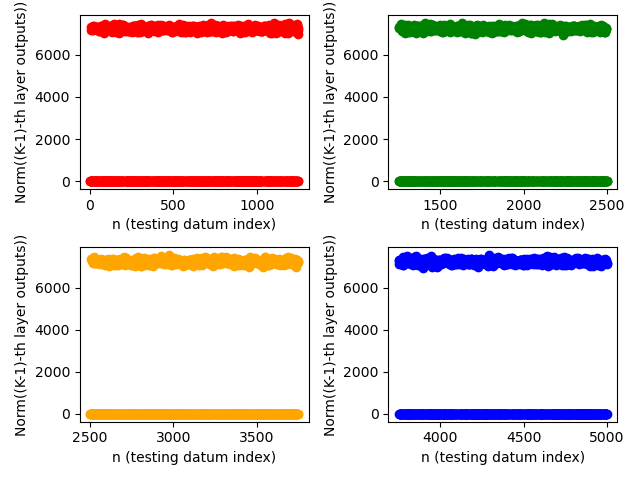}
	\caption{A scatter plot of $\| \bm{y}_{K-1,n}^{(T)} \|$ versus $n$ under High-SNR-T, $(K, H)=(20,100)$, and testing at 35 dB: the computed testing $P_e$ at 35 dB is $P_e=1.0$.}
	\label{fig: ReLU_FNN_high_SNR_Norm_of_the_penultimate_layer_output_vectors_K_20_H_100}
\end{figure}

\begin{figure}[htb!]
	\centering
	\includegraphics[scale=0.45]{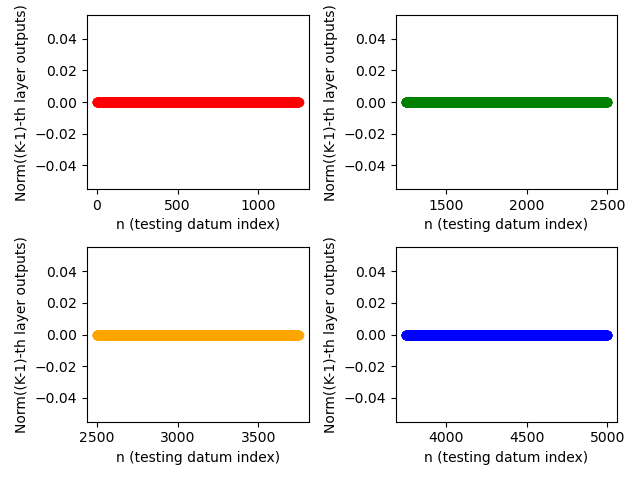}
	\caption{A scatter plot of $\| \bm{y}_{K-1,n}^{(T)} \|$ versus $n$ under Low-SNR-T, $(K, H)=(20,100)$, and testing at 35 dB: the computed testing $P_e$ at 35 dB is $P_e=1.0$.}
	\label{fig: ReLU_FNN_low_SNR_Norm_of_the_penultimate_layer_output_vectors_K_20_H_100}
\end{figure}

We now proceed to deep and wider ReLU FNN-based BPSK detectors. Figs. \ref{fig: ReLU_FNN_all_SNRs_Norm_of_the_penultimate_layer_output_vectors_K_20_H_100}-\ref{fig: ReLU_FNN_low_SNR_Norm_of_the_penultimate_layer_output_vectors_K_20_H_100} show the scatter plots of $\| \bm{y}_{K-1,n}^{(T)} \|$ versus $n$ subject to All-SNR-T, High-SNR-T,\linebreak and Low-SNR-T, respectively, for $(K, H)=(20,100)$. These plots fail to validate Theorem \ref{thm_performance_limits_ReLU_FNN_based_detector} since the manifested $P_e$ is always 1.0, affirming some testing results that cannot be interpreted using our theory.      

 \begin{table*}[!t]
 	\centering
 	\begin{tabular}{ | l | c | c | c | c | c | c | c | }
 		\hline
 		\hspace{1mm}SNR & \multicolumn{1}{ c | }{Optimal} & \multicolumn{3}{ c | }{FNN-based BPSK detector}  & \multicolumn{3}{ c | }{ FNN-based BPSK detector}  \\
 		$[$in dB$]$ & \multicolumn{1}{ c | }{detector} & \multicolumn{3}{ c | }{$(K, H)=(3,3)$}  & \multicolumn{3}{ c | }{$(K, H)=(3,9)$}  \\
 		\hline
 		$\gamma_b$ $[$dB$]$ & $\mathcal{Q}(\sqrt{2\gamma_b})$ & All-SNR-T & High-SNR-T & Low-SNR-T & All-SNR-T & High-SNR-T & Low-SNR-T    \\  \hline  
 		0 & 0.0786 & 0.54196 & 0.60334  & 0.54232001  & 0.54198  & 0.57132  & 0.54192001      \\  \hline
 		5 & 0.0060 & 0.50231999  & 0.51339999  & 0.50235999  & 0.50231999  & 0.50687999  & 0.50229999       \\  \hline
 		10 & $3.8721\times 10^{-6}$ & 0.50075999 & 0.50075999  & 0.50075999  & 0.50075999  & 0.50075999  & 0.50075999      \\  \hline
 		15 & $9.1240\times 10^{-16}$ & 0.50338   & 0.50338  & 0.50338  & 0.50338  & 0.50338   &  0.50338     \\  \hline
 		20 & $1.0442 \times 10^{-45} $ & 0.49835998  & 0.49835998 & 0.49835998  & 0.49835998   & 0.49835998  & 0.49835998      \\  \hline
 		25 & $7.3070 \times 10^{-140} $ & 0.49908  & 0.49908  & 0.49908  &  0.49908  & 0.49908  &  0.49908     \\  \hline
 		30 & 0 & 0.49677998  & 0.49677998  & 0.49677998  & 0.49677998 & 0.49677998 & 0.49677998     \\  \hline
 		35 & 0 & 0.50049999  & 0.50049999  & 0.50049999  & 0.50049999  & 0.50049999  & 0.50049999      \\  \hline
 	\end{tabular}   \\ [4mm]
 	\caption{Numerical results I on bit error rate manifested by optimal BPSK detector and BPSK detectors based on FNNs with ReLU and Tanh activation.}
 	\label{computer_experiment_results_I_for_FNNs}
 \end{table*}
 \begin{table*}[htb!]
 	\centering
 	\begin{tabular}{ | l | c | c | c | c | c | c | c | }
 		\hline
 		\hspace{1mm}SNR & \multicolumn{1}{ c | }{Optimal} & \multicolumn{3}{ c | }{ FNN-based BPSK detector}  & \multicolumn{3}{ c | }{ FNN-based BPSK detector}    \\
 		$[$in dB$]$  &  \multicolumn{1}{ c | }{detector} & \multicolumn{3}{ c | }{$(K, H)=(4, 8)$}  & \multicolumn{3}{ c | }{$(K, H)=(7, 7)$}    \\
 		\hline
 		$\gamma_b$ $[$dB$]$ & $\mathcal{Q}(\sqrt{2\gamma_b})$ & All-SNR-T & High-SNR-T &Low-SNR-T  & All-SNR-T  & High-SNR-T & Low-SNR-T     \\  \hline  
 		0 & 0.0786 & 0.54194  & 0.58012 & 0.54179999 & 0.54192001  & 0.60139999  & 0.54192001         \\  \hline
 		5 & 0.0060 & 0.50229999 & 0.50854  & 0.50226  & 0.50229999  & 0.51298001  & 0.50229999         \\  \hline
 		10 & $3.8721\times 10^{-6}$ & 0.50075999  & 0.50075999 & 0.50075999  & 0.50075999  & 0.50075999 & 0.50075999       \\  \hline
 		15 & $9.1240\times 10^{-16}$ & 0.50338  & 0.50338   & 0.50338  & 0.50338   & 0.50338  & 0.50338       \\  \hline
 		20 & $1.0442 \times 10^{-45} $ & 0.49835998  & 0.49835998  & 0.49835998  & 0.49835998  & 0.49835998  & 0.49835998        \\  \hline
 		25 & $7.3070 \times 10^{-140} $ & 0.49908  & 0.49908  & 0.49908  & 0.49908  & 0.49908  & 0.49908       \\  \hline
 		30 & 0 & 0.49677998 & 0.49677998 & 0.49677998  & 0.49677998  & 0.49677998  & 0.49677998       \\  \hline
 		35 & 0 & 0.50049999  & 0.50049999  & 0.50049999 & 0.50049999  & 0.50049999  & 0.50049999        \\  \hline
 	\end{tabular}   \\ [4mm]
 	\caption{Numerical results II on bit error rate manifested by optimal BPSK detector and BPSK detectors based on FNNs with ReLU and Tanh activation.}
 	\label{computer_experiment_results_II_for_FNNs}
 \end{table*}

\begin{figure}[htb!]
	\centering
	\includegraphics[scale=0.45]{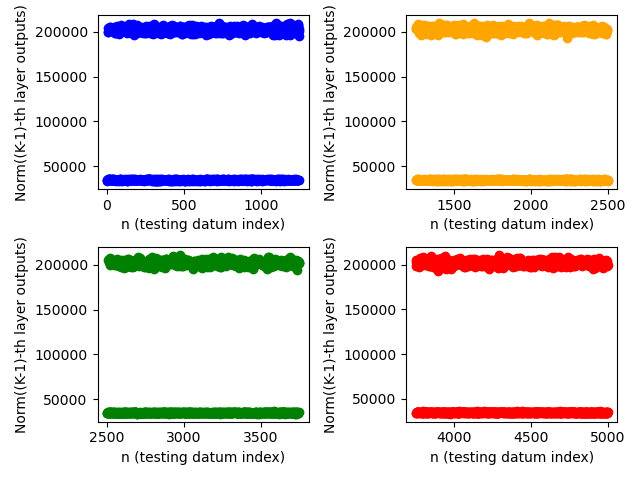}
	\caption{A scatter plot of $\| \check{\bm{y}}_{K-1,n}^{(T)} \|$ versus $n$ under All-SNR-T, $(K, H)=(8,8)$, and testing at 35 dB: the computed testing $\check{P}_e$ at 35 dB is $\check{P}_e=0.50049999$.}
	\label{fig: FNN_all_SNRs_Norm_of_the_penultimate_layer_output_vectors_K_8_H_8}
\end{figure}

\begin{figure}[htb!]
	\centering
	\includegraphics[scale=0.45]{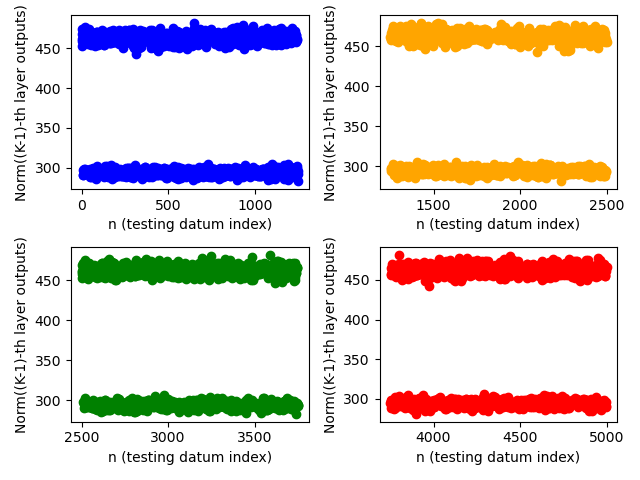}
	\caption{A scatter plot of $\| \check{\bm{y}}_{K-1,n}^{(T)} \|$ versus $n$ under High-SNR-T, $(K, H)=(8,8)$, and testing at 35 dB: the computed testing $\check{P}_e$ at 35 dB is $\check{P}_e=0.50049999$.}
	\label{fig: FNN_high_SNR_Norm_of_the_penultimate_layer_output_vectors_K_8_H_8}
\end{figure} 

\begin{figure}[htb!]
	\centering
	\includegraphics[scale=0.45]{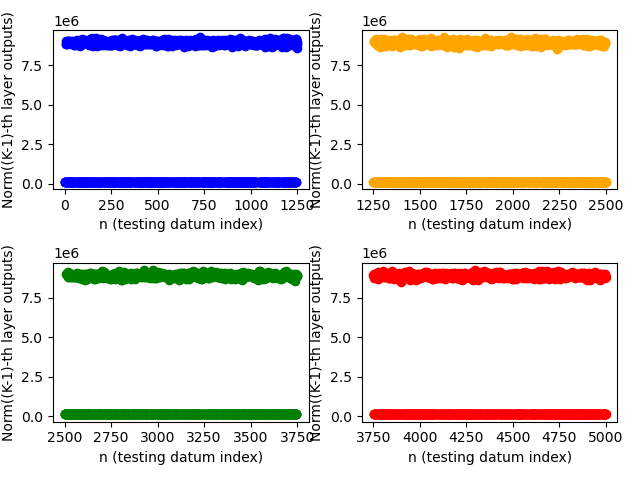}
	\caption{A scatter plot of $\| \check{\bm{y}}_{K-1,n}^{(T)} \|$ versus $n$ under Low-SNR-T, $(K, H)=(8,8)$, and testing at 35 dB: the computed testing $\check{P}_e$ at 35 dB is $\check{P}_e=0.50049999$.}
	\label{fig: FNN_low_SNR_Norm_of_the_penultimate_layer_output_vectors_K_8_H_8}
\end{figure}

\begin{figure}[htb!]
	\centering
	\includegraphics[scale=0.45]{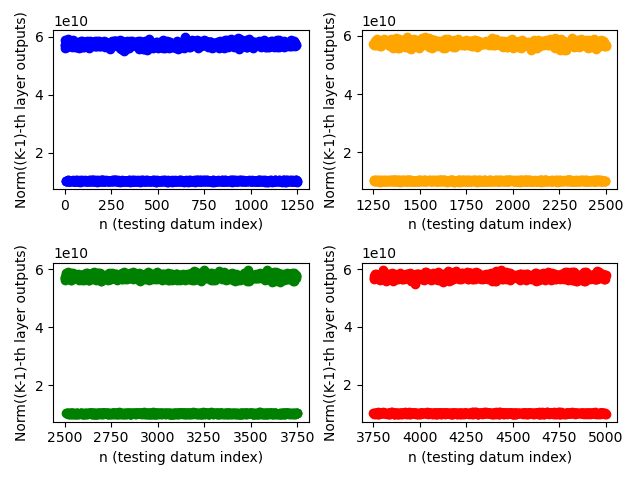}
	\caption{A scatter plot of $\| \check{\bm{y}}_{K-1,n}^{(T)} \|$ versus $n$ under All-SNR-T, $(K, H)=(20,100)$, and testing at 35 dB: the computed testing $\check{P}_e$ at 35 dB is $\check{P}_e=0.50049999$.}
	\label{fig: FNN_all_SNRs_Norm_of_the_penultimate_layer_output_vectors_K_20_H_100}
\end{figure}
\begin{figure}[htb!]
	\centering
	\includegraphics[scale=0.45]{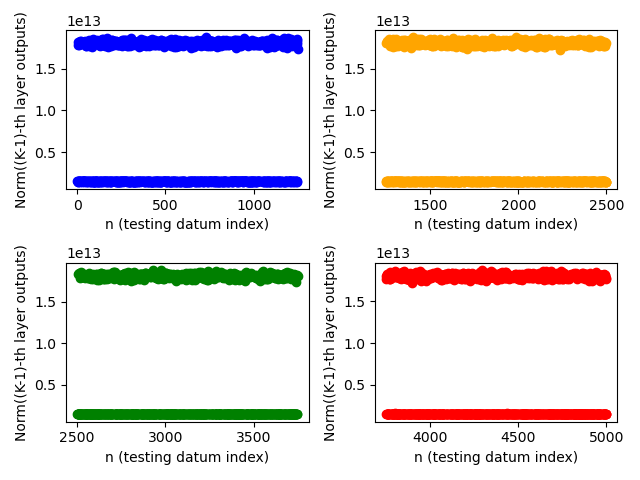}
	\caption{A scatter plot of $\| \check{\bm{y}}_{K-1,n}^{(T)} \|$ versus $n$ under High-SNR-T, $(K, H)=(20,100)$, and testing at 35 dB: the computed testing $\check{P}_e$ at 35 dB is $\check{P}_e=1.0$.}
	\label{fig: FNN_high_SNR_Norm_of_the_penultimate_layer_output_vectors_K_20_H_100}
\end{figure} 
\begin{figure}[htb!]
	\centering
	\includegraphics[scale=0.45]{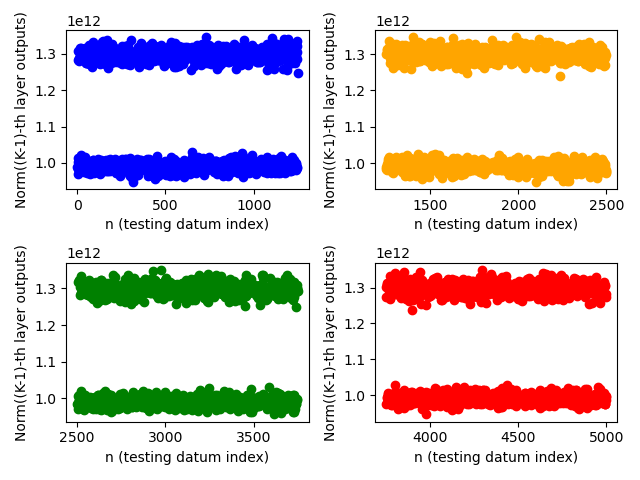}
	\caption{A scatter plot of $\| \check{\bm{y}}_{K-1,n}^{(T)} \|$ versus $n$ under Low-SNR-T, $(K, H)=(20,100)$, and testing at 35 dB: the computed testing $\check{P}_e$ at 35 dB is $\check{P}_e=0.50049999$.}
	\label{fig: FNN_low_SNR_Norm_of_the_penultimate_layer_output_vectors_K_20_H_100}
\end{figure} 

\subsection{Results for FNNs with ReLU and Tanh Activation}
\label{subsec: main_results_FNNs}
Employing the (hyper)parameters of \cite[Table VII]{arXiv_Getu_Fundamental_Limits'23}, we generated the testing and theory validation results -- presented in Sections \ref{subsubsec: testing_results_FNNs} and \ref{subsubsec: theory_validation_results_FNNs}, respectively -- for BPSK detectors based on FNNs with ReLU and Tanh activation.    
\subsubsection{Testing Results}
\label{subsubsec: testing_results_FNNs}
The numerical results for BPSK detectors based on FNNs with ReLU and Tanh activation are provided in Tables \ref{computer_experiment_results_I_for_FNNs} and \ref{computer_experiment_results_II_for_FNNs}. As seen in those tables, BPSK detectors based on FNNs -- with ReLU and Tanh activation -- trained with hinge loss and \texttt{Adam} fail to learn optimum BPSK detection for any SNR.      

\subsubsection{Theory Validation Results}
\label{subsubsec: theory_validation_results_FNNs}
As to the BPSK detectors based on shallow FNNs with ReLU and Tanh activation, Figs. \ref{fig: FNN_all_SNRs_Norm_of_the_penultimate_layer_output_vectors_K_8_H_8}-\ref{fig: FNN_low_SNR_Norm_of_the_penultimate_layer_output_vectors_K_8_H_8} show the scatter plots of $\| \check{\bm{y}}_{K-1,n}^{(T)} \|$ versus $n$ subject to All-SNR-T, High-SNR-T, and Low-SNR-T, respectively, for $(K, H)=(8,8)$. Figs. \ref{fig: FNN_all_SNRs_Norm_of_the_penultimate_layer_output_vectors_K_8_H_8}\&\ref{fig: FNN_low_SNR_Norm_of_the_penultimate_layer_output_vectors_K_8_H_8} validate Theorem \ref{thm_performance_limits_FNN_based_detector_with_tanh} since the exhibited $\check{P}_e$ equals 0.5 and $\| \check{\bm{y}}_{K-1,n}^{(T)} \|$ tends to zero or infinity. However, Theorem \ref{thm_performance_limits_FNN_based_detector_with_tanh}\linebreak is not verified by Fig. \ref{fig: FNN_high_SNR_Norm_of_the_penultimate_layer_output_vectors_K_8_H_8}, which is among the results that cannot be interpreted by our developed theory.   

As for the BPSK detectors based on deep and wider FNNs with ReLU and Tanh activation, Figs. \ref{fig: FNN_all_SNRs_Norm_of_the_penultimate_layer_output_vectors_K_20_H_100}-\ref{fig: FNN_low_SNR_Norm_of_the_penultimate_layer_output_vectors_K_20_H_100} show the scatter plots of $\|\check{\bm{y}}_{K-1,n}^{(T)} \|$ versus $n$ subject to All-SNR-T, High-SNR-T, and Low-SNR-T, respectively, for $(K, H)=(20,100)$. Figs. \ref{fig: FNN_all_SNRs_Norm_of_the_penultimate_layer_output_vectors_K_20_H_100}\&\ref{fig: FNN_low_SNR_Norm_of_the_penultimate_layer_output_vectors_K_20_H_100} validate Theorem \ref{thm_performance_limits_FNN_based_detector_with_tanh} because the manifested $\check{P}_e$ equals 0.5 and $\| \check{\bm{y}}_{K-1,n}^{(T)} \|$ tends to zero or infinity. Nevertheless, Theorem \ref{thm_performance_limits_FNN_based_detector_with_tanh} is not validated by Fig. \ref{fig: FNN_high_SNR_Norm_of_the_penultimate_layer_output_vectors_K_20_H_100}, which is among the testing results uninterpretable by our theory.

\section{Conclusions and Research Outlook}
\label{sec: Conc_remarks_and_outlook}
Despite DL's broad impacts and successes, a comprehensive insight on why and how DL is empirically successful in astronomically many AI/ML problems remains fundamentally elusive. In this vein, no fundamental work has managed yet to quantify the testing performance limits of DL-based classifiers trained with cross-entropy loss or hinge loss. To overcome this limitation in part, we developed a theory for the asymptotic testing performance limits of deep ReLU FNNs and deep FNNs with ReLU and Tanh activation. Our theory is an intuitive one that interprets some results of our extensive computer experiments, especially as the corresponding FNNs get deeper and wider. 

Important research outlook is developing a fundamental theory on the non-asymptotic performance limits of binary classifiers based on $i)$ deep ReLU FNNs and $ii)$ deep FNNs with ReLU and Tanh activation. What is also important is determining the fundamental testing performance limits of DL-based multi-level classifiers trained using cross-entropy loss, which undoubtedly entails more complex decision regions/boundaries.

	\ifCLASSOPTIONcompsoc
	\section*{Acknowledgments}
	\else
	\section*{Acknowledgment}
	\fi
	
	T. M. Getu acknowledges Prof. Ali H. Sayed (EPFL, Switzerland) for fruitful discussions and criticisms; the U.S. Department of Commerce and its agency NIST for a computational support and a prior funding support.

	\ifCLASSOPTIONcaptionsoff
	\newpage
	\fi

	

\begin{IEEEbiography}[{\includegraphics[width=1.1in,height=1.25in,clip,keepaspectratio]{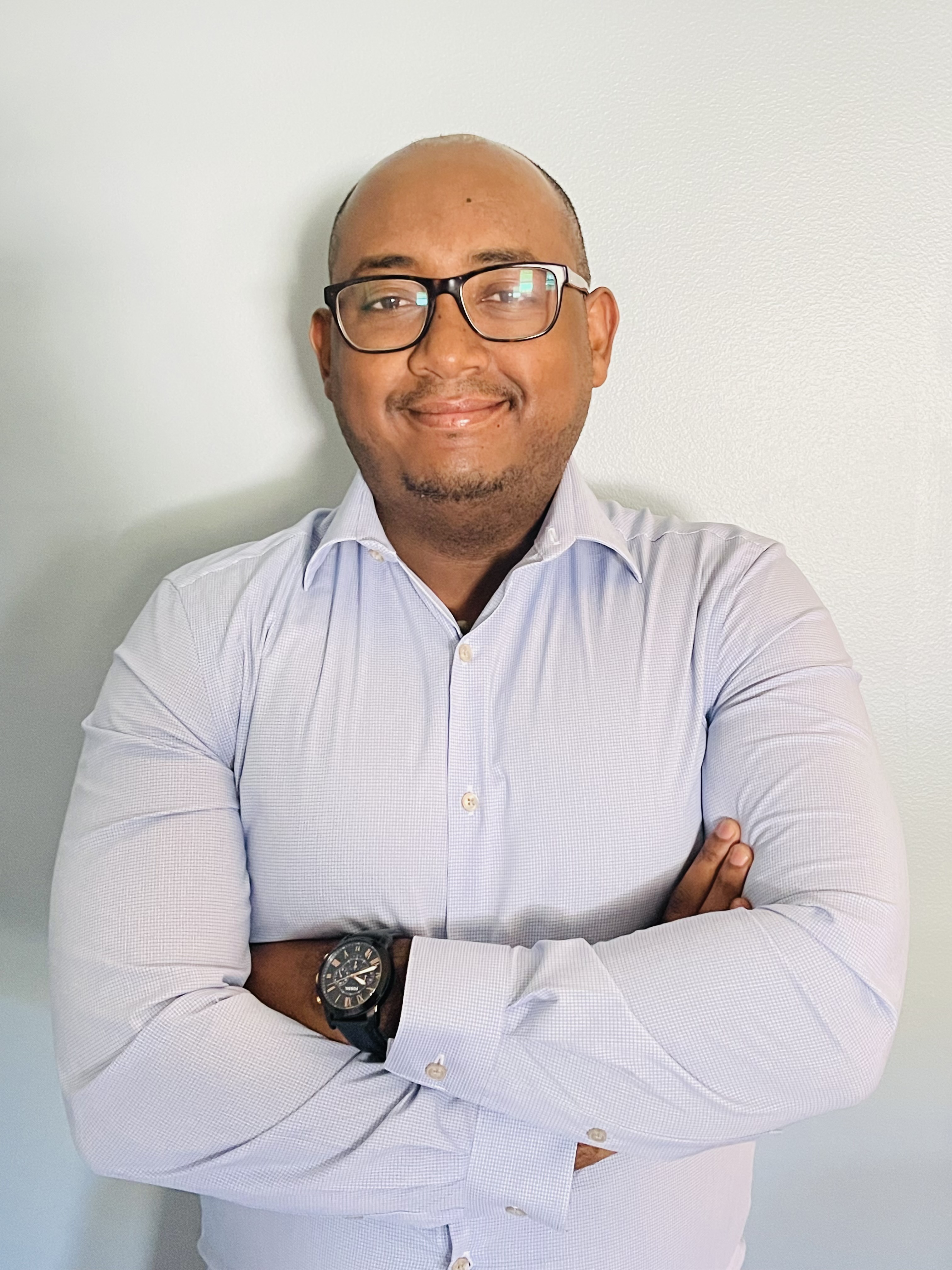}}]{\textbf{Tilahun M. Getu}} (M'19) earned the Ph.D. degree (with highest honor) in electrical engineering from the \'Ecole de Technologie Sup\'erieure (\'ETS), Montreal, QC, Canada in 2019. He is currently a Post-doctoral Fellow with the \'ETS. His transdisciplinary fundamental research interests span the numerous fields of classical and quantum \textbf{S}cience, \textbf{T}echnology, \textbf{E}ngineering, and \textbf{M}athematics (\textbf{STEM}) at the nexus of communications, signal processing, and networking (all types); intelligence (both artificial and natural); robotics; computing; security; optimization; high-dimensional statistics; and high-dimensional causal inference.   
	
Dr. Getu has received several awards, including the 2019 ÉTS Board of Director’s Doctoral Excellence Award in recognition of his Ph.D. dissertation selected as the 2019 \'ETS all-university best Ph.D. dissertation.
\end{IEEEbiography}

\begin{IEEEbiography}[{\includegraphics[width=1.1in,height=1.25in,clip,keepaspectratio]{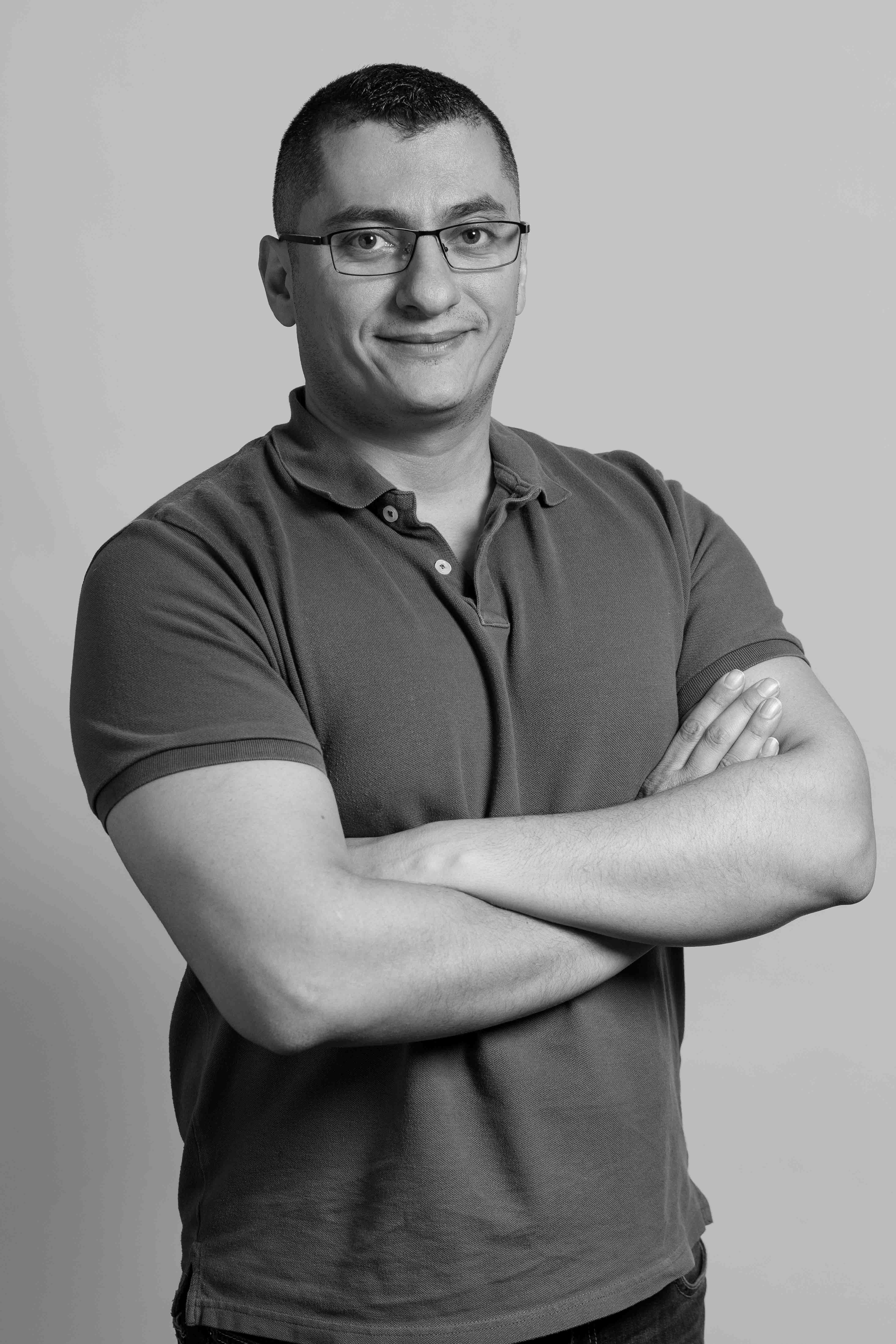}}]{\textbf{Georges Kaddoum}} (M'11--SM'20) is a professor and Tier 2 Canada Research Chair with the École de Technologie Supérieure (ÉTS), Université du Québec, Montréal, Canada. He is also a Faculty Fellow in the Cyber Security Systems and Applied AI Research Center at Lebanese American University. His recent research activities cover 5G/6G networks, tactical communications, resource allocations, and security.  Dr. Kaddoum has received many prestigious national and international awards in recognition of his outstanding research outcomes. Currently, Prof. Kaddoum serves as an Area Editor for the IEEE Transactions on Machine Learning in Communications and Networking and an Associate Editor for IEEE Transactions on Information Forensics and Security, and IEEE Transactions on Communications. 
\end{IEEEbiography}

\begin{IEEEbiography}[{\includegraphics[width=1.1in,height=1.25in,clip,keepaspectratio]{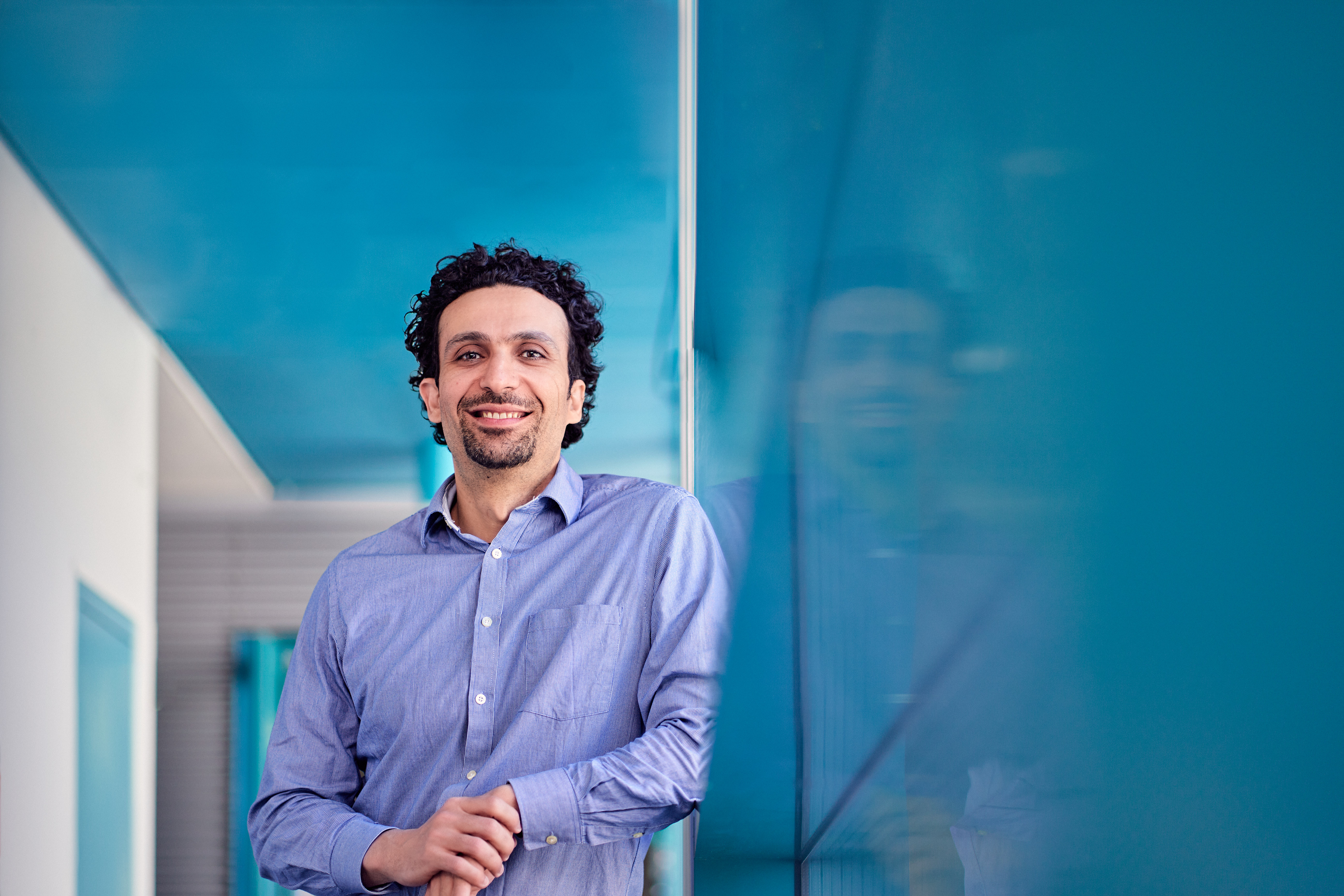}}]{\textbf{Mehdi Bennis}} (F’20) is a full tenured Professor at the Centre for Wireless Communications, University of Oulu, Finland and head of the \textbf{I}ntelligent \textbf{CO}nnectivity and \textbf{N}etworks/Systems Group (\textbf{ICON}). His main research interests are in radio resource management, game theory and distributed AI in 5G/6G networks. He has published more than 200 research papers in international conferences, journals and book chapters. He has been the recipient of several prestigious awards. Dr. Bennis is an editor of IEEE TCOM and Specialty Chief Editor for Data Science for Communications in the Frontiers in Communications and Networks journal.
\end{IEEEbiography}

\clearpage
\includepdf[pages={1-11}]{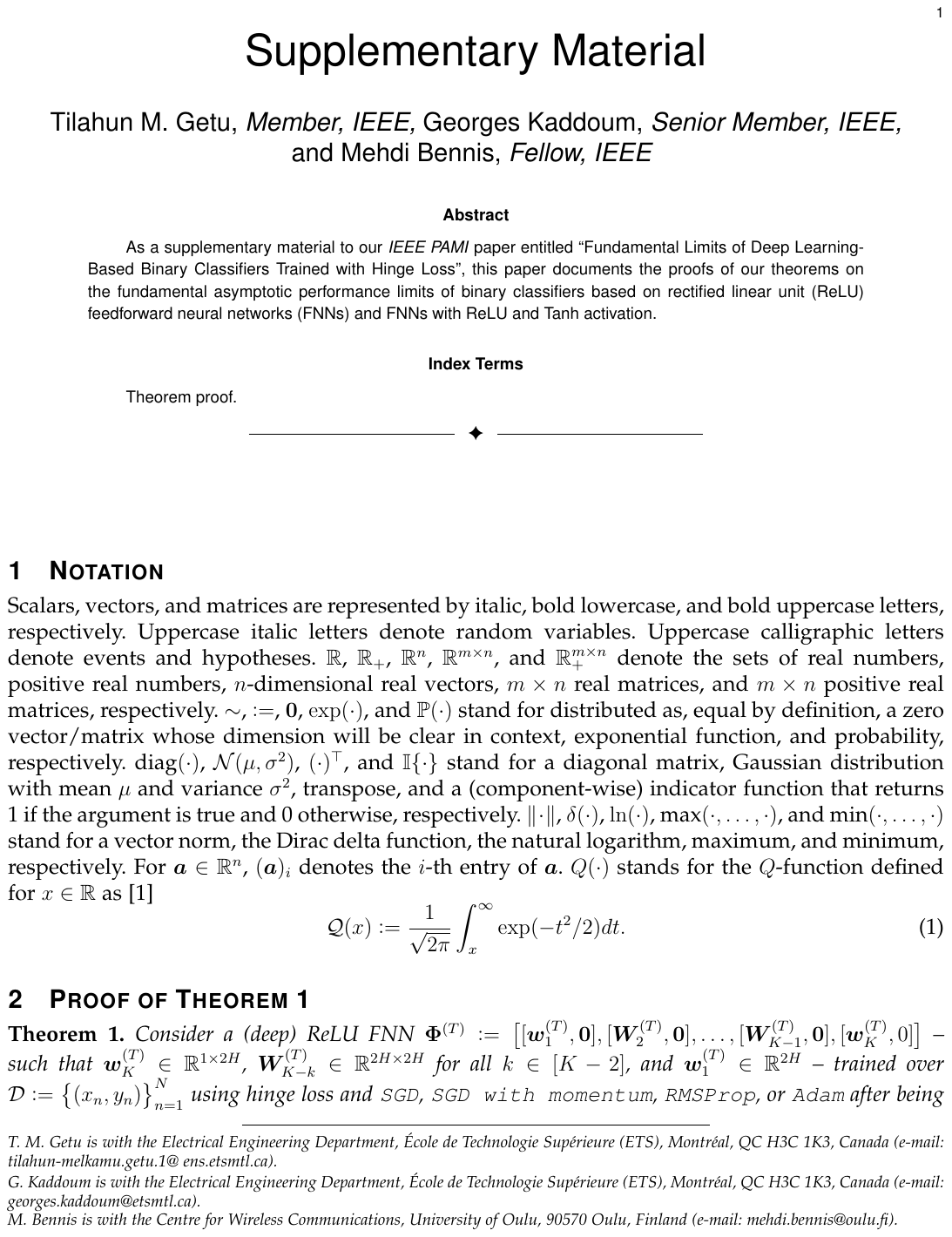}

\end{document}